\documentclass[11pt]{article}
\usepackage{bm}
\usepackage{geometry}
\usepackage{fullpage}
\usepackage{url}
\usepackage[english]{babel}
\usepackage{graphicx}
\usepackage{color}
\usepackage{rotating}
\usepackage{tabularx}
\usepackage{natbib}
\usepackage{amsmath}
\usepackage{comment}
\includecomment{comment}
\usepackage{kbordermatrix}

\usepackage[unicode=true,pdfusetitle,
 bookmarks=true,bookmarksnumbered=true,bookmarksopen=true,bookmarksopenlevel=2,
 breaklinks=true,pdfborder={0 0 1},backref=false,colorlinks=true,citecolor=blue,urlcolor=blue,linkcolor=blue]
 {hyperref}

\usepackage{amssymb}
\usepackage{subcaption}
\usepackage{tipa}

\title{Modeling morphology with Linear Discriminative Learning: \\
considerations and design choices}
\author{Maria Heitmeier, Yu-Ying Chuang, and R. Harald Baayen\\Eberhard-Karls Universität T\"{u}bingen }
\date{November 2021}

\begin{document}

\maketitle

\begin{abstract}

\noindent
This study addresses a series of methodological questions that arise when modeling inflectional morphology with Linear Discriminative Learning.  Taking the semi-productive German noun system as example, we illustrate how decisions made about the representation of form and meaning influence model performance.  We clarify that for modeling frequency effects in learning, it is essential to make use of incremental learning rather than the end-state of learning.  We also discuss how the model can be set up to approximate the learning of inflected words in context.  In addition, we illustrate how in this approach the wug task can be modeled.  The model provides an excellent memory for known words, but appropriately shows more limited performance for unseen data, in line with the semi-productivity of German noun inflection and generalization performance of native German speakers.

\end{abstract}

\section{Introduction} 

Computational models of morphology fall into two broad classes.  The first class addresses the question of how to produce a morphologically complex word given a morphologically related form (often a stem, or an identifier of a stem or lexeme) and a set of inflectional or derivational features.  We refer to these models as form-oriented models.  The second class comprises models seeking to understand the relation between words' forms and their meanings. We refer to these models as meaning-oriented models. 

Prominent form-oriented models comprise Analogical Modeling of Language \citep[AML;][]{Skousen:89,Skousen:2002} and Memory Based Learning \citep[MBL;][]{Daelemans:VandenBosch:2005}, which are nearest-neighbor classifiers.  Input to these models are tables with observations (words) in rows, and factorial predictors and a factorial response in columns.  The response specifies an observation's outcome class (e.g., an allomorph), and the model is given the task to predict the outcome classes from the other predictor variables (for allomorphy,  specifications of words' phonological make-up).  Predictions are based on sets of nearest neighbors, serving as constrained exemplar sets for generalization. These models have clarified morphological phenomena ranging from the allomorphy of the Dutch diminutive \citep{Daelemans:95} to stress assignment in English \citep{arndt2011towards}. 

\citet{Ernestus:Baayen:2003} compared the performance of the MBL, AML, and Generalised Linear (GLM) models, as well as a recursive partitioning tree \citep{Breiman:84}, on the task of predicting whether word-final obstruents in Dutch alternate with respect to their voicing.  They observed similar performance across all models, with the best performance, surprisingly,  for the only  parameter-free model, AML.  Their results suggest that the quantitative structure of morphological datasets may be straightforward to discover for any reasonably decent classifier. The model proposed by \citet{belth2021greedy} is a recent example of a classifier based on recursive partitioning. 

Minimum Generalization Learning \citep[MGL;][]{Albright:Hayes:2003} offers an algorithm for rule induction \citep[for comparison with nearest neighbor methods, see][]{KeuleersCogPsy:2007}. The model finds rules by an iterative process of minimal generalization that combines specific rules into ever more general rules.  Each rule comes with a measure of prediction accuracy, and the rule with the highest accuracy is selected for predicting a word's form.

All models discussed thus far are exemplar-based, in the sense that the input to any of these models consists of a table with exemplars, exemplar features selected on the basis of domain knowledge, and a categorical response variable specifying targeted morphological form changes.  In other words, all these models are classifiers that absolve the analyst from hand-engineering lexical entries,  rules or constraints operating on these lexical entries, and theoretical constructs such as inflectional classes.  In this respect, they differ fundamentally from the second group of the following computational methods. 

The DATR language \citep{Evans:Gazdar:96} defines non-monotonic inheritance networks for knowledge representation.  This language is optimized for removing redundancy from lexical descriptions.  A DATR model requires the analyst to set up lexical entries that specify information about, for instance, inflectional class, gender, the forms of exponents, and various kinds of phonological information.  The lexicon is designed in such a way that the network is kept as small as possible, while still allowing the model, through its mechanism of inheritance,  to correctly predict all inflected variants.   Realizational morphology \citep[RM;][]{stump2001imt} sets up rules for realizing bundles of inflectional and lexical features in phonological form. This theory can also be defined as a formal language (a finite-state transducer) that provides mappings from underlying representations onto their corresponding surface forms and vice versa \citep{karttunen2003computing}.  The Gradual Learning Algorithm \citep[GLA;][]{Boersma:98,Boersma:Hayes:2001} works within the framework of optimality theory \citep{prince2008optimality}. The algorithm is initialized with a set of  constraints and gradually learns an optimal constraint ranking by incrementally moving through the training data, and upgrading or downgrading constraints. 

The third group of form-oriented computational models comprises connectionist models. The past-tense model of \citet{Rumelhart:McClelland:86b} was trained to produce English past-tense forms given the corresponding present-tense form. An early enhancement of this model was proposed by \citet{MacWhinney:Leinbach:91}, for an overview of the many follow-up models, see \citet{kirov2018recurrent}. Kirov and Cotterell proposed a sequence-to-sequence deep learning network, the Encoder-Decoder (ED) learner, that they argue does not suffer from the drawbacks noted by \citet{Pinker:Prince:88} for the original paste-tense model. \citet{malouf2017abstractive} introduced a recurrent deep learning model trained to predict upcoming segments, showing that this model has high accuracy for predicting paradigm forms given the lexeme and the inflectional specifications of the desired paradigm cell.  

In summary, the class of form-oriented models comprises three subsets: statistical classifiers (AML, MBL, GLM, recursive partitioning),  generators based on linguistic knowledge engineering (DATR, RM, GLA),  and connectionist models (paste-tense model, ED learner).  The models just referenced presuppose that when speakers use a morphologically complex form, this form is derived on the fly from its underlying form. The sole exception is the model of \citet{malouf2017abstractive}, which takes the lexeme and its inflectional features as point of departure.  As pointed out by \citet{Blevins:2016}, the focus on how to create one form from another has its origin in pedagogical grammars, which face the task of clarifying to a second language learner how to create inflected variants. Unsurprisingly, applications within natural language processing also have need of systems that can generate inflected and derived words.  

However, it is far from self-evident that native speakers of English would create past-tense forms from present-tense forms.    Meaning-oriented models argue that in comprehension, the listener or reader can go straight from the auditory or visual input to the intended meaning, without having to go through a pipeline requiring initial identification of underlying forms and exponents. Likewise, speakers are argued to start from meaning, and realize this meaning directly in written or spoken form. 

The class of meaning-oriented models comprises both symbolic and subsymbolic models.  The symbolic models of \citet{Levelt:Roelofs:Meyer:1999} and \citet{dell1986spreading} implement a form of realizational morphology.   Concepts and inflectional features activate stems and exponents, which are subsequently combined into syllables.  Both models hold that the production of morphologically complex words is a compositional process
in which units are assembled together and ordered for articulation at various hierarchically ordered levels. These models have been worked out only for English, and to our knowledge have not been applied to languages with richer morphological systems.

The subsymbolic model of \citet{Harm:Seidenberg:2004} sets up multi-layer networks between orthographic, phonological, and semantic units.  No attempt is made to define morphemes, stems, or exponents.  To the extent that such units have any reality, they are assumed to arise, statistically, at the hidden layers.  \citet{Mirkovic:MacDonald:Seidenberg:2005} argue for Serbian that  that gender is an emergent property of the network that arises from statistical regularities governing both words' forms and their meanings \citep[see][for discussion of semantic motivations for gender systems]{corbett1991gender}. The model for auditory comprehension of \citet{Gaskell:WMW:1997} uses a three-layer recurrent network to map speech input onto distributed semantic representations, again without attempting to isolate units such as phonemes or morphemes.   

The naive discrimination learning (NDL) model proposed by \citet{Baayen:Milin:Filipovic:Hendrix:Marelli:2011} represents words' forms sub-symbolically, but words' meanings symbolically.   The modeling set-up that we discuss in the remainder of this study, that of linear discriminative learning \citep[LDL,][]{Baayen:Chuang:Shafei:Blevins:2019}, replaces the symbolic representation of word meaning in NDL by sub-symbolic representations building on distributional semantics \citep[][]{Landauer:Dumais:1997,mikolov2013distributed}.  

LDL is an implementation of Word and Paradigm Morphology \citep{Matthews:1974,Blevins:2016}. Sublexical units such as stems and exponents play no role.  Semantic representations in LDL, however, are analytical: the semantic vector (word embedding, i.e. a distributed representation of meaning) of an inflected word is constructed from the semantic vector of the lexeme  and the semantic vectors of the pertinent inflectional functions.  Both NDL and LDL make use of the simplest possible networks: networks with only input and output layers, and no hidden layers.

At this point, the distinction made by \citet{breiman2001statistical} between statistical models and machine learning is relevant.  Statistical models aim to provide insight into the mechanisms that  generate the data.  Machine learning, on the other hand, aims to optimize prediction accuracy, and it is not an issue whether or not  the algorithms are interpretable. LDL is much closer to statistical modeling than to the black boxes of machine learning. All input and output representations can be set up in a theoretically transparent way \citep{Baayen:Chuang:Shafei:Blevins:2019}.  Furthermore, because LDL implements multivariate multiple regression, its mathematical properties are well-understood.  Importantly, modeling results do not depend on the choice of hyper-parameters (e.g., the numbers of LSTM layers and LSTM units), instead, they are completely determined by the representations chosen by the analyst.

The goals of this study are, first, to clarify how such choices of representation affect LDL model performance; second, to illustrate what can be achieved simply with multivariate multiple regression; and third, to call attention to the kind of problems that are encountered when word meaning is integrated into morphology.  Our working example is the comprehension and production of German nouns.  In what follows, we first introduce the German noun system, and review models that have been proposed for German nouns.  We then introduce LDL, after which we present a systematic overview of modeling choices, covering the representation of form, the representation of meaning, and the learning algorithm (incremental learning versus the regression `end-state of learning' solution).

\section{German noun morphology}\label{sec:german}

The German noun system is both highly irregular and semi-productive, featuring three different genders, two numbers and four cases. In this section we will give an overview over this system, show where irregularity and semi-productivity arise, and which (non-computational) models have been employed to account for it.

\begin{table}[tb!]
    \centering
    \begin{tabular}{l|l|r} \hline
        Plural class & Example & Type frequency\\
        \hline
        \textit{-(e)n} & Tasse $\rightarrow$ Tassen `cup(s)' & 56.5\%\\
        \textit{(uml+)-e} &  Tag $\rightarrow$ Tage `day(s)' & \\
        & Topf $\rightarrow$ Töpfe `pot(s)` & 23.9\%\\
        \textit{(uml+)-er} & Brett $\rightarrow$ Bretter `board(s)' & \\ 
         & Glas $\rightarrow$ Gläser `glass(es)' & 2.3\%\\
        \textit{(uml+)-}$0$ & Daumen $\rightarrow$ Daumen `thumb(s)' & \\
        & Apfel $\rightarrow$ Äpfel `apple(s)' & 13.3\% \\
        \textit{-s} & Kamera $\rightarrow$ Kameras `camera(s)' & 2.6\%\\ \hline
    \end{tabular}
    \caption{Plural classes of German nouns (relative frequencies from \citet{gaeta2008deutsche}). Most of the classes can appear with both masculine and neuter nouns. Feminine nouns belong mostly to the \textit{-(e)n} class (97\%).}
    \label{tab:plural_classes}
\end{table}

Plural forms are marked with one of four suffixes (\textit{-(e)n}, \textit{-er}, \textit{-e}, \textit{-s}) or without adding a suffix ($-0$; a ``zero'' morpheme \citep[][p. 306]{kopcke1988schemas}), three of which can pair with stem vowel fronting (e.g. \textit{a} (\textipa{/a/}) $\rightarrow$ \textit{ä} (\textipa{/E/})) \citep[e.g.][]{kopcke1988schemas} (Table~\ref{tab:plural_classes}). There are additional suffixes which usually apply to words with foreign origin, such as \textit{-i} (e.g. \textit{Cello} $\rightarrow$ \textit{Celli}, `cellos') \citep{cahill1999german}.  \citet{cahill1999german} sub-categorize the nouns into 11 classes, based on whether singulars have a different suffix than plurals (\textit{Album} $\rightarrow$ \textit{Alben}, `albums').  \citet{nakisa1996defaults} distinguish between no less than 60 inflection classes. No plural class is prevalent overall \citep{kopcke1988schemas}, and it is impossible to fully predict plural class from gender, syntax, phonology or semantics \citep{kopcke1988schemas, cahill1999german, trommer2021subsegmental}.  
Further complications arise when case is taken into account.  German has four cases: \textit{nominative, genitive, dative}, and \textit{accusative}, which are marked with two exponents (applied additional to the plural markers):  \textit{-(e)n} and \textit{-(e)s} \citep{Schulz1981}. Case forms are also not fully predictable from gender, phonology or meaning. Since many forms do not receive a separate marker, the system has been described as  ``degenerate'' \citep[][p. 245]{bierwisch2018syntactic} (see Table~\ref{tab:declension}). German speakers do, however, get additional disambiguing information from the definite and indefinite articles which accompany nouns and likewise encode gender, number and case Table~\ref{tab:declension} shows the definite articles for all genders. Additionally, there are indefinite articles available for singular forms which also express case in their endings (e.g. Gen. sg. m./n./f. \textit{eines}, Dat. sg. m./n. \textit{einem}, Dat. sg. f. \textit{einer}).

\begin{table}[tb!]
    \centering
    \begin{tabular}{l|l|l|l|r} \hline
        case \& number & masculin I & masculin II & neutral & feminin  \\
        \hline
        Nom. sg. & der Freund         & der Mensch         & das Kind         & die Mutter \\
        Gen. sg. & des Freund{\bf es} & des Mensch{\bf en} & des Kind{\bf es} & der Mutter \\
        Dat. sg. & dem Freund         & dem Mensch{\bf en} & dem Kind         & der Mutter \\
        Acc. sg. & den Freund         & den Mensch{\bf en} & das Kind         & die Mutter \\
        Nom. pl. & die Freund{\bf e}  & die Mensch{\bf en} & die Kind{\bf er} & die Mütter \\
        Gen. pl. & der Freund{\bf e}  & der Mensch{\bf en} & der Kind{\bf er} & der Mütter\\
        Dat. pl. & den Freund{\bf en} & den Mensch{\bf en} & den Kind{\bf ern}& den Mütter{\bf n} \\
        Acc. pl. & die Freund{\bf e}  & die Mensch{\bf en} & die Kind{\bf er} & die Mütter\\ \hline
    \end{tabular}
    \caption{German noun declension.  Plural endings vary with declension class.  Table adapted from \citet[][p. 105]{Schulz1981}.}
    \label{tab:declension}
\end{table}

Unsurprisingly, it has been the subject of a long-standing debate whether a distinction between regular and irregular nouns is useful for German (the debate has mostly focused on the formation of the nominative plural which we accordingly also focus on here).  It is also unsurprising that the system shows limited productivity.  Several so-called `wug' studies, where participants are asked to inflect nonce words,  have clarified that German native speakers struggle with predicting unseen plurals. \citet{kopcke1988schemas, zaretsky2013acquisition, mccurdy2020inflecting} reported high variability across speakers with respect to the plural forms produced.   \citet{kopcke1988schemas} took this as evidence for a ``modified schema model'' of German noun inflection, arguing that plural forms are generated based not only on a speaker's experience with the German noun system, but also on the ``cue validity'' of the plural markers. For example, \textit{-(e)n} is a good cue for plurality, as it does not occur with many singular forms.  By contrast,  \textit{-er} has low cue validity for plurality, as it occurs with many singulars.  

\citet{kopcke1988schemas} also observed that \textit{-s} is used slightly more in his wug experiments than would be expected from corpus data. \citet{marcus1995german} and \citet{clahsen1999lexical}  therefore argued that \textit{-s} serves as the regular default plural marker in German, contrasting with all other plural markers that are described as irregular. Others, however, have argued that an \textit{-s} default rule does not provide any additional explanatory value \citep{nakisa1996defaults, zaretsky2015no, behrens1999and, indefrey1999some}. 

Despite the irregularity and variability in the system, some sub-regularities within the German noun system have also been pointed out \citep{wunderlich1999german, wiese1999default}. For instance, \citet[p.7f.]{wunderlich1999german} reports a set of rules that German nouns adhere to, which can be overridden on an item-by-item basis through `lexical storage'. For example, he notes that 
\begin{quote}
    a. Masculines ending in schwa are weakly inflected (and thus also have n-plurals). \\
    b. Non-umlauting feminines have an n-plural.\\
    c. Non-feminines ending in a consonant have a      \textipa{@}-plural. [\ldots]\\
    e. All atypical nouns have an s-plural. [\ldots]
\end{quote}
He also allows for semantics to co-determine class membership.  For instance, masculine animate nouns show a tendency to belong to the \textit{-n} plural class \citep[see also][]{gaeta2008deutsche}. 
A further remarkable aspect of the German noun system, especially for second language learners, is that whereas it is remarkably difficult to learn to produce the proper case-inflected forms, understanding these forms in context is straightforward. 

In the light of these considerations, the challenges for computational modeling of German noun inflection, specifically from a cognitive perspective, are the following:
\begin{enumerate}
    \item to construct a memory for a highly irregular, ``degenerate'', semi-productive system,
    \item to ensure that this memory shows some moderate productivity for novel forms, but with all the uncertainties that characterize the generalization capacities of German native speakers,
    \item to furthermore ensure that the performance of the mappings from form to meaning, and from meaning to form, within the framework of the discriminative lexicon \citep{Baayen:Chuang:Shafei:Blevins:2019}, are properly asymmetric with respect to comprehension and production accuracy \citep[see also][]{Chuang:Bell:Banke:Baayen:2020}.
\end{enumerate}

\subsection{Computational models for German nouns}

The complexity of the German declension system has inspired many  computational models. The DATR model of \citet{cahill1999german} belongs to the class of generating models based on linguistic knowledge engineering.  It assigns lexemes to carefully designed hierarchically ordered declension classes. Each class inherits the properties from classes further up in the hierarchy, but will override some of these properties.  This model provides a successful and succint formal model for German noun declension. Other models from this class include GERTWOL which is based on finite-state operations \citep{haapalainen1994gertwol}, as well as the model of \citet{trommer2021subsegmental} which draws on Optimality Theory (OT) and likewise requires careful hand-crafting and constraint ranking (but does not currently have a computational implementation).  

\citet{belth2021greedy} propose a statistical classifier based on recursive partitioning, with as response variable the morphological change required to transform a singular into a plural, and as predictors the final segments of the lexeme, number, and case. At each node, nouns are divided by their features, with one branch comprising the most frequent plural ending (which will inevitably include some nouns with a different plural ending, labelled as exceptions), and with the other branch including the remainder of the nouns. Each leaf node of the resulting tree is said to be productive if a criterion for node homogeneity is met.  An older model, also a classifier building up rules inductively was developed 20 years earlier by \citet{Albright:Hayes:2003}. 

Connectionist models for the German noun system include a model using a simple recurrent network \citep{Goebel:Indefrey:2000},  and a deep learning model implementing a sequence-to-sequence encoder-decoder \citep{mccurdy2020inflecting}.  The latter model takes letter-based representations of German nouns in their nominative singular form as input, together with information on the grammatical gender of the noun. The model is given the task to produce the corresponding nominative plural form.  The model learned the task with high accuracy on held out data (close to 90\%), but was more locked in on the `correct' forms compared to native speakers, who in a wug task showed substantially more variability in their choices. 

The models discussed above also differ with respect to how they generate predictions for novel nouns.  The sequence-to-sequence deep learning model of \citep{mccurdy2020inflecting} can do so relatively easily, straight from a word's form and its gender specification but its inner workings are not immediately interpretable \citep[though recent work has started to gain some insights, see e.g.][for syntactic structure in deep learning]{linzen2021syntactic}. By contrast, the linguistically more transparent DATR model can only generate a novel word's inflectional variants once this word has been assigned to an inflectional class.  This may to some extent be possible given its principal parts \citep{finkel2007principal}, but clearly requires additional mechanisms to be in place.

In what follows, we introduce the LDL model. LDL is a model of human lexical processing, with all its limitations and constraints, rather than an optimized computational system for generating (or understanding) morphologically complex words. It implements a simple linear mapping between form and meaning, where form is represented as a binary vector of sublexical cues, and meaning is represented in a distributed fashion.

By applying LDL to the modeling of the German noun system (including its case forms), we address a question that has thus far not been addressed computationally, namely the incorporation of semantics.  Semantic subregularities in the German noun system have been noted by several authors \citep[e.g.][]{wunderlich1999german, gaeta2008deutsche}, and although deep learning models can be set up that incorporate semantics \citep[see, e.g.,][]{malouf2017abstractive}, LDL by design must take semantics into account. 

The next section introduces the LDL model. The following sections proceed with an overview of the many modeling decisions that have to be made.  An important part of this overview is devoted to moving beyond the modeling of isolated words, as words come into their own only in context \citep{Elman:2009dinosaur}, and case labels do not correspond to contentful semantics, but instead are summary devices for syntactic distribution classes \citep{Blevins:2016, Baayen:Chuang:Shafei:Blevins:2019}.

\section{Linear Discriminative Learning}\label{sec:ldl}

LDL is the computational engine of the discriminative lexicon model (DLM) proposed by \citet{Baayen:Chuang:Shafei:Blevins:2019}.   The DLM implements mappings between form and meaning for both reading and listening, and mappings from meaning to form for production. It also allows for multiple routes operating in parallel.  For reading in English, for instance, it sets up a direct route from form to meaning, in combination with an indirect route from visual input to a phonological representation that in turn is mapped onto the semantics \citep[cf.][]{coltheart1993models}. In what follows, we restrict ourselves to the mappings from form onto meaning (comprehension) and from meaning onto form (production).
Mappings can be obtained either with trial-to-trial learning, or by estimating the end-state of learning.  In the former case, the model implements incremental regression using the learning rule of \citet{Widrow:Hoff:1960}; in the latter case, it implements multivariate multiple linear regression, which is mathematically equivalent to a simple network with input units, output units, no hidden layers, and simple summation of incoming activation without using thresholding or squashing functions. 

Each word form of interest is represented by a set of cues. For example, \textit{wordform1} might feature the cues \texttt{cue1}, \texttt{cue2} and \texttt{cue3}, while \textit{wordform2} could be marked by \texttt{cue1}, \texttt{cue4} and \texttt{cue5}.  We can thus express a word form as a binary vector, where 1 denotes presence and 0 absence. This information is coded in the cue matrix $\mathbf{C}$: \\

$$
\renewcommand{\kbldelim}{(}% Left delimiter
\renewcommand{\kbrdelim}{)}% Right delimiter
    \mathbf{C} = \kbordermatrix{ & \mathtt{cue1} & \mathtt{cue2} & \mathtt{cue3} & \mathtt{cue4} & \mathtt{cue5} \\ 
    \text{wordform1} & 1 & 1 & 1 & 0 & 0 \\
    \text{wordform2} & 1 & 0 & 0 & 1 & 1 \\ }
$$
\ \\

\noindent
Words' meanings are also represented by numeric vectors. The dimensions of these vectors can have a discrete interpretation, or have a latent interpretation (see Section~\ref{sec:semantics} below for detailed discussion). In the following example, \textit{wordform1} has strong negative support for semantic dimensions \texttt{S3} and \texttt{S5}, while \textit{wordform2} has strong positive support for \texttt{S4} and \texttt{S5}. This information is brought together in a semantic matrix $\mathbf{S}$: \\

$$
\kbalignrighttrue
\renewcommand{\kbldelim}{(}% Left delimiter
\renewcommand{\kbrdelim}{)}% Right delimiter
\renewcommand{\kbrowstyle}{\displaystyle}
\renewcommand{\kbcolstyle}{\displaystyle}
    \mathbf{S} = \kbordermatrix{ ~ & \texttt{S1} & \texttt{S2} & \texttt{S3} & \texttt{S4} & \texttt{S5} \cr \text{wordform} 1 & 0.1 & 0.004 & -1.95 & 0.03 & -0.54 \cr \text{wordform} 2 & -0.49 & -0.32 & 0.03 & 1.06 & 0.98 \cr }
$$
\ \\

Comprehension and production in LDL are modelled by means of simple linear mappings from the form matrix $\bm{C}$ to the semantic matrix $\bf{S}$, and vice versa.  These mappings specify how strongly input nodes are associated with output nodes.  The weight matrix for a given mapping can be obtained in two ways. First, using the mathematics of multivariate multiple regression, a comprehension weight matrix $\mathbf{F}$ is obtained by solving $\mathbf{F}$ from
$$
\mathbf{S} = \mathbf{C}\cdot\mathbf{F},
$$
and a production weight matrix $\mathbf{G}$ is obtained by solving $\mathbf{G}$ from
$$
\mathbf{C} = \mathbf{S}\cdot\mathbf{G}.
$$
As for linear regression modeling, the predicted row vectors are approximate.  Borrowing notation from statistics, we write
$$
\hat{\mathbf{S}} = \mathbf{C} \cdot \mathbf{F}
$$
for predicted semantic vectors (row vectors of $\hat{\bm{S}}$), and
$$
\hat{\mathbf{C}} = \mathbf{S} \cdot \mathbf{G}
$$
for predicted form vectors (row vectors of $\hat{\mathbf{C}}$).  

These equations amount to estimating multiple outcomes from multiple variables, which in statistics is referred to as multivariate multiple regression. In simple linear regression, a single value $y$ is estimated from a value $x$ via an intercept $\beta_0$ and a weighing of $x$ with scalar $\beta_1$: \begin{equation}
    \hat{y} = \beta_0 + \beta_1 x
\end{equation} which can easily be expanded to estimating $y$ from a vector $\mathbf{x}$ (multiple linear regression), using a vector of beta coefficients $\beta_i \in \bm{\beta}$ to weigh each value $x_i \in \mathbf{x}$: \begin{equation}
    \hat{y} = \beta_0 +  x_1\beta_1 +  x_2\beta_2 + ... +  x_n\beta_n
\end{equation}

Finally, to estimate a vector $\mathbf{y}$ from a vector $\mathbf{x}$ (multivariate multiple regression), we need an entire matrix of beta coefficients $\beta_{ij} \in \mathbf{B}$. A single value $y_{i} \in \mathbf{y}$ is then estimated via  
\begin{equation}
    \hat{y}_{i} = \beta_{0i}  + x_{1} \beta_{1i} + x_{2} \beta_{2i} + \ldots + x_{p} \beta_{pi}
\end{equation}
Thus, estimating the mappings $\mathbf{F}$ and $\mathbf{G}$ in LDL amounts to computing the coefficients matrix $\mathbf{B}$ for mappings from $\mathbf{C}$ to $\mathbf{S}$ and vice versa. As such, each value in a predicted semantic vector $\hat{\mathbf{s}}$ (form vector $\hat{\mathbf{c}}$) is a linear combination (i.e. weighted sum) of the values in the corresponding form vector $\mathbf{c}$ (semantic vector $\mathbf{s}$) it is predicted from. This means that LDL is mathematically highly constrained: it cannot handle non-linearities that even shallow connectionist models  \citep{goldsmith2006learning} can take in their stride. Nevertheless, we have found that these simple linear mappings result in high accuracies \citep[e.g.][]{Baayen:Chuang:Blevins:2018, Baayen:Chuang:Shafei:Blevins:2019} suggesting that morphological systems are surprisingly simple.
Cases where model predictions are less precise due to the limitations of linearity become indicative of learning bottlenecks.

Furthermore, note that estimating the mappings $\mathbf{F}$ and $\mathbf{G}$ using the matrix algebra of multivariate multiple regression provides optimal estimates, in the least squares sense, of the connection weights (or equivalently, beta coefficients) for datasets that are type-based, in the sense that each pair of row vectors $\bm{c}$ of $\bm{C}$ and $\bm{s}$ of $\bm{S}$ is unique.  Having multiple instances of the same pair of row vectors in the dataset does not make sense, as it renders the input completely singular and does not add any further information.  Thus, models based on the regression estimates of $\mathbf{F}$ and $\mathbf{G}$ are comparable to type-based models such as AML, MBL, MGL, and models using recursive partitioning.

Making the estimates of the mappings sensitive to frequency of use requires incremental learning, updating weights after each word token that is presented for learning.  Incremental learning is implemented using the learning rule of \citet{Widrow:Hoff:1960,milin2020keeping}, which defines the matrix $\bm{W}^{t+1}$ with updated weights at time $t+1$ as the weight matrix $\bm{W}^t$ at time $t$, modified as follows:
$$
\bm{W}^{t+1} = \bm{W}^t + \bm{c} \cdot (\bm{o}^T - \bm{c}^T \cdot \bm{W}^t) \cdot \eta,
$$ 
where $\bm{c}$ is the current cue (vector), $\bm{o}$ the current outcome vector, and $\eta$ the learning rate. Conceptually, this means that after each newly encountered word token, the weight matrix is changed such that the next time that the same cue vector has to be mapped onto its associated outcome vector, it will be slightly closer to the target outcome vector than it was before.  The learning rule of Widrow-Hoff implements incremental regression.  As the number of times that a model is trained again and again on a training set increases (training epochs), the network's weights will converge to the matrix of beta coefficients obtained by approaching the estimation problem with multivariate multiple regression \citep[see, e.g.][]{Chuang:Bell:Banke:Baayen:2020,ShafaeiBajestan:2021, Evert:Arppe:2015}.  As a consequence, the regression-based estimates pertain to the `end-state of learning', at which the data have been worked through infinitely many times.  Unsurprisingly, effects of frequency and order of learning are not reflected in model predictions based on the regression estimates.  Such effects do emerge with incremental learning (see Section~\ref{sec:kind_of_learning}).   

Comprehension accuracy for a given word $\omega$ is assessed by comparing its predicted semantic vector $\hat{s}_\omega$ with all gold standard semantic vectors in $\bm{S}$ (the creation of gold standard semantic vectors will be described in subsequent sections), using either the cosine similarity measure or the Pearson correlation $r$.  In what follows, we use $r$, and select as the meaning that is recognized that gold standard row vector $\bm{s}_\text{max}$ of $\bm{S}$ that shows the highest correlation with $\hat{s}_\omega$.  If $\bm{s}_\text{max}$ is the targeted semantic vector, the model's prediction is classified as correct, otherwise, it is taken to be incorrect.

For the modeling of production, a supplementary algorithm is required for constructing actual word forms.  The predicted vectors $\hat{\bm{c}}$ provide information about the amount of support that form cues receive from the semantics.  However, information about the amount of support received by the full set of cues does not provide information about the order in which a subset of these cues have to be woven together into actual words.   Algorithms that construct words from form cues make use of the insight that when form cues are defined as n-grams ($n > 1$), the cues contain implicit information about order.  For instance, for digraph cues, cues {\tt ab} and {\tt bc} can be combined into the string {\tt abc}, whereas cues {\tt ab} and {\tt cd} cannot be merged.  Therefore, when n-grams are used as cues, directed edges can be set up in a graph with n-grams as vertices, for any pair of n-grams that properly overlap.  A word form is uniquely defined by a path is such a graph starting with an initial n-gram (starting with an initial word edge symbol, typically a \# is used) and ending at a final n-gram (ending with \#).  This raises the question of how to find  word paths in the graph.  This is accomplished by first discarding n-grams with low support from the semantics below a threshold $\theta$\footnote{This is a simple cut-off point for n-grams with low support, not to be confused with thresholds as often used in deep learning.}, then calculating all possible remaining paths, and finally selecting for articulation that path for which the corresponding predicted semantic vector (obtained by mapping its corresponding cue vector $\bm{c}$ onto $\bm{s}$ using comprehension matrix $\bm{F}$) best matches the semantic vector that is the target for articulation. This implements  `synthesis by analysis', see \citet{Baayen:Chuang:Shafei:Blevins:2019} and \citet{Baayen:Chuang:Blevins:2018} for further details and theoretical motivation. For a discussion of the cognitive plausibility of this method see \citet{chuang2020estonian}.

The first algorithm that was used to enumerate possible paths made use of a shortest-paths algorithm from graph theory.  This works well for small datasets, but becomes prohibitively expensive for large datasets.  The {\bf JudiLing} package \citep{Luo:Chuang:Baayen:2021} offers a new algorithm that scales up better.  This algorithm is first trained to predict,  from either the $\hat{\bm{C}}$ or the $\bm{S}$ matrix, for each possible word position, which cues are best supported at that position.  All possible paths with the top $k$ best-supported cues are then calculated, and subjected to synthesis by analysis. Details about this algorithm, implemented in julia in the {\bf JudiLing} package as the function {\tt learn\_paths} can be found in \citep{luo2021judiling}.  The {\tt learn\_paths} function is used throughout the remainder of this study.  A word form is judged to be produced correctly when it exactly matches the targeted word form.

\section{Modelling considerations} \label{sec:modelling}

\begin{figure}[tb!]
    \centering
    \includegraphics[width=\textwidth]{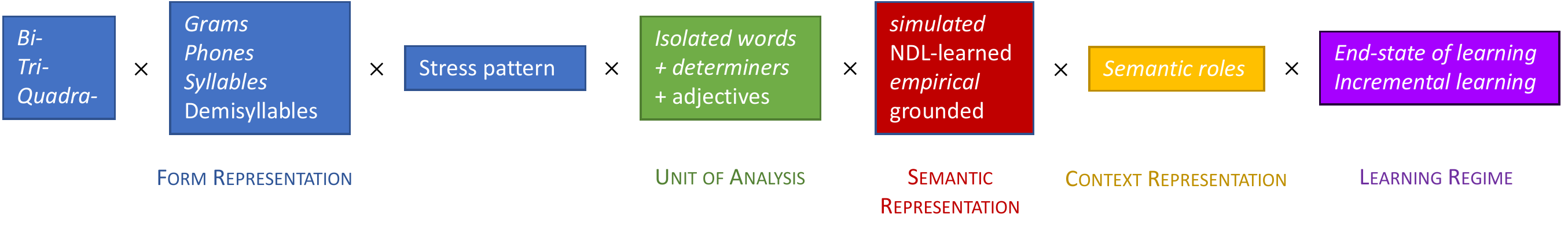}
    \caption{Options when modelling a language's morphology with LDL. Examples with options in \textit{italics} are discussed in the present study.}
    \label{fig:choices}
\end{figure}

When modelling a language's morphology within the framework of the DLM, the analyst is faced with a range of  choices, illustrated in Figure~\ref{fig:choices}.  From left to right, choices are listed for representing form, for the unit of analysis, for the representation of semantics, for the handling of context, and for the learning regime.

With respect to form representations, the kind of n-gram has to be selected,  setting $n$, deciding on phonological or orthographic grams, and specifying how stress or lexical tone are represented.  With respect to the unit of analysis, the analyst has to decide whether to model isolated words, or words in phrasal contexts.  A third set of choices concerns what semantic representations to use: simulated representations, or word embeddings such as {\em word2vec} \citep{mikolov2013distributed}, or grounded vectors \citep{shahmohammadi2021learning}.  A further set of choices for languages with case concerns how to handle case labels, as these typically refer to syntactic distribution classes rather than contentful inflectional features \citep{Blevins:2016}.  Finally, a selection needs to be made with respect to whether incremental learning is used, or instead the end-state of learning using regression-based estimation.  In what follows, we illustrate several of these choice points using examples addressing the German noun system. 

\begin{table}[tb!]
        \centering
        \begin{tabular}{l|l|l|l|l|l|r} \hline
            word form & pronunciation & lemma & case & number & frequency & gender \\
            \hline
            Aal & \texttt{al} & Aal & nominative & singular & 29 & m \\
            Aal &\texttt{al} & Aal & dative & singular & 29& m \\
            Aal &\texttt{al} & Aal & accusative & singular &29 & m \\
            Aale &\texttt{al@} & Aal & nominative & plural &34 & m \\
            Aale &\texttt{al@} & Aal & genitive & plural &34 & m \\
            Aalen &\texttt{al@n} & Aal & dative & plural & 17& m \\
            Aalen &\texttt{al@n} & Aal & accusative & plural &17 & m \\ \hline
        \end{tabular}
        \caption{Representation of the paradigm for \textit{Aal} `eel' in our dataset. Genitive singular (\textit{Aals}) is not included as it has a frequency of 0 in CELEX}.
        \label{tab:paradigm}
\end{table}

The dataset on German noun inflection that we use for our worked examples was compiled as follows. First, we extracted all monomorphemic nouns and their inflections with a frequency of at least 1 from CELEX \citep{baayen1995celex}, resulting in a dataset of about 6,000 word forms. Of these we retained the 5,486 word forms for which we could retrieve grammatical gender from Wiktionary, thus including word forms of 2,732 different lemmas. The resulting data was expanded such that each attested word form was listed once for each possible paradigm cell it could belong to.  For instance, \textit{Aal} (`eel') is listed once as singular nominative, once as dative and once as accusative  (Table~\ref{tab:paradigm}). This resulted in a dataset with 18,147 entries, with word form frequencies ranging from 1 to 5,828, (M log frequency 2.56, SD  1.77). Word forms are represented in their DISC notation, which represents German phones with single characters\footnote{Data and code are available in the supplementary materials at \url{https://osf.io/zrw2v/}.}. Table~\ref{tab:paradigm} clarifies that there are many homophones. As a consequence, the actual number of distinct word forms in our dataset is only 5,486, which amounts to on average about two word forms per lemma.

There are many ways in which model performance can be evaluated.  First, we may be interested in how well the model performs as a memory.  How well does the model learn to understand and produce words it has encountered before?  Note that because the model is not a list of forms, this is not a trivial question.  For evaluation of the model as a memory, we consider its performance on the training data (henceforth \texttt{train}).  Second, we may be interested in the extent to which the memory is productive. Does it generalize so that new forms can be understood or produced?  Above, we observed that the German noun system is semi-regular, and that German native speakers are unsure about what the proper plural is of words they have not encountered before \citep{mccurdy2020inflecting}.  If our modeling approach properly mirrors human limitations on generalization from data with only partial regularities, evaluation on unseen, held-out data of German should not be perfect.   At this point, however, several issues arise that require careful thought.   

For one, from  the perspective of the linguistic system, it seems unreasonable to assume that any held-out form can be properly produced (or understood) if some of the  principal parts \citep{finkel2007principal} of the lexeme are missing in the training data.  In what follows, we will make the simplifying assumption that under cross-validation with sufficient training data, this situation will not arise.   

A further question that arises is how to evaluate held-out words that have homophones in the training data.  Such homophones present novel combinations of a form vector (shared with another data point in the training data) and a semantic vector (not attested for this form in the training data). We may want to impose a strict evaluation criterion requiring that the model gets the semantic vector exactly right.  However, when presented with a homophone in isolation, a human listener cannot predict which of a potentially large set of paradigm cells is the targeted one (the problem of modeling words in isolation).  We may therefore want to use a lenient evaluation criterion for comprehension according to which comprehension is judged to be accurate when the predicted semantic vector $\hat{\bm{s}}$ is associated with one of a homophonic word's possible semantic interpretations.  Yet a further possible evaluation metric is to see how well the model performs on words with forms that have not been encountered in the training data. These possibilities are summarized in Table~\ref{tab:eval}. Below, in section~\ref{sec:careful_split}, we consider further complications that can arise in the context of testing the model on unseen forms.

\begin{table}[tb!]
\centering
\begin{tabular}{llll} \hline
evaluation type &  \\ \hline
 simple         &  blind evaluation of all held-out data    &    & {\tt val\_all}             \\ \hline
 nuanced        &  evaluation on novel forms only  &             & {\tt val\_newform}         \\ \hline
                &  evaluation on homophones        &  strict     & {\tt val\_strict}  \\
                &                                  &  lenient    & {\tt val\_lenient} \\ \hline
\end{tabular}
\caption{Types of model evaluation} 
\label{tab:eval}
\end{table}

For evaluating the productivity of the model, we split the full dataset into 80\% training data and 20\% validation data, with 14,518 and 3,629 word forms respectively. In the validation data, 3309 forms are also present in the training data (i.e. homophones), and 320 are new forms. Among the 320 new forms, 8 have novel lemmas that are absent in the training data. Since it is unrealistic to expect the model to understand or produce inflected forms of completely new words, these 8 words are excluded from the validation dataset for new forms, although they are taken into consideration when calculating the overall accuracy for the validation data. The same training and validation data are used for all the simulations reported below, unless indicated otherwise.

\subsection{Representing words' forms}\label{sec:form}

Decisions about how to represent words' forms depend on the modality that is to be modelled. For auditory comprehension, \citet{Arnold:Tomaschek:Lopez:Sering:Baayen:2017} and \citet{ShafaeiBajestan:2021} explore ways in which form vectors can be derived from the audio signal.  Instead of using low-level audio features, one can also use more abstract symbolic representations such as phone n-grams.\footnote{Other work \citep[e.g.][]{Joanisse:Seidenberg:1999} has used slot-coding for representing phonology, but we do not think that this representation is optimal, since, for example, we are not sure how prefixation is to be modeled without hand-engineering \citep[details in][]{Heitmeier:Baayen:2020}.}  For visual word recognition, one may use letter n-grams, or, as lower-level visual cues, for instance, features derived from histograms of oriented gradients \citep{Dalal:Triggs:2005,Linke:Baayen:2017}.  In what follows, we use binary vectors indicating the presence or absence of phonological phone or syllable n-grams.

\subsubsection{Phone-based representations}

Sublexical phone cues can be of different granularity, such as biphones and triphones.  For the word {\em Aale} (pronunciation {\tt al@}), the biphone cues are {\tt \#a}, {\tt al}, {\tt l@}, and {\tt @\#}, and the triphone cues are {\tt \#al}, {\tt al@}, and {\tt l@\#}.  The number of unique cues (and hence the dimensionality of the form vectors) increases as granularity decreases. For the present dataset, there are 931 unique biphone cues, but 4,656 triphone cues.  For quadraphones, there are no less than 9,068 unique cues. Although model performance tends to become better with more unique cues, we also run the risk of overfitting. That is, the model does not generalize and thus performs worse on validation data. The choice of granularity therefore determines the balance of having a precise memory on the one hand and a productive memory on the other hand.  In the simulation examples with n-phones that follow, we made use of simulated semantic vectors.  Details on the different kinds of semantic vectors that can be used are presented in Section \ref{sec:simulated_vec}. 

\begin{table}[tb!]
{\small
    \centering
    \begin{tabular}{l|c|c|cc|c|c|cc} \hline
                   & \multicolumn{4}{c}{comprehension} & \multicolumn{4}{|c}{production} \\ \cline{2-9}
                   & train & val\_all & val\_lenient & val\_newform & train & val\_all & val\_lenient & val\_newform  \\ \hline
        biphone    &  22\% & 16\% & 17\% & 8\% & 48\% & 31\% & 33\% & 12\% \\
       triphone    &  93\% & 88\% & 92\% & 51\% & 84\% & 64\% & 68\% & 21\% \\
       quadraphone &  97\% & 93\% & 97\% & 53\% & 91\% & 67\% & 73\% & 11\%  \\ \hline
       bisyllable  &  99\% & 93\% & 99\% & 20\% & 95\% & 63\% & 69\% & 0.3\%  \\
       \hline
       {\em word2vec} &  87\% & 72\% & 79\% & 0.3\% & 97\% & 88\% & 94\% & 25\%  \\ \hline
    \end{tabular}
}
    \caption{Comprehension and production accuracy for train and validation datasets, with biphones, triphones, quadraphones, and bisyllables as cues. For the first four rows, we used simulated semantic vectors. For the last row, cues are triphones, and semantic vectors are {\em word2vec} embeddings (discussed in Section \ref{sec:emp_vec}). For the \texttt{learn\_paths} algorithm, the threshold $\theta$ was set to 0.05, 0.008, 0.005, 0.005, and 0.008 respectively.}
    \label{tab:form_res}
\end{table}

Accuracy for n-phones is presented in the first three rows of Table \ref{tab:form_res}.  For the training data, comprehension accuracy is high with both triphones and quadraphones. For biphones, the small number of unique cues clearly does not offer sufficient discriminatory power to distinguish word meanings.  Under strict evaluation,  unsurprisingly given the large number of homophones in German noun paradigms, comprehension accuracy plummets substantially to 8\%, 33\%, and 35\% for biphone, triphone, and quadraphone models respectively. Given that there is no way to tell the meanings of homophones apart without further contextual information, we do not provide further details for strict evaluation.  However, in Section~\ref{sec:def_article} we will address the problem of homophony by incorporating further contextual information into the model.

With regards to model accuracy for validation data, we see that overall accuracy ({\tt val\_all}) is quite low for biphones, while it remains high for both triphones and quadraphones. Closer inspection reveals that this high accuracy is mainly contributed by homophones ({\tt val\_lenient}). Since these forms are already present in the training data, a high comprehension accuracy under lenient evaluation is unsurprising. As for unseen forms (i.e., {\tt val\_newform}), quadraphones perform slightly better than triphones.

Production accuracy, presented in the right half of Table \ref{tab:form_res}, is highly sensitive to the threshold $\theta$ used by the {\tt learn\_paths} algorithm.   Given that usually only a relatively small number of cues receive strong support from a given meaning, we therefore set the threshold such that the algorithm does not need to take into account large numbers of irrelevant cues.  Depending on the form and meaning representations selected,  some fine-tuning is generally required to obtain a threshold value that optimally balances both accuracy and computation time.  Once the threshold is fine-tuned for the training data, the same threshold is used for the validation data.

Production accuracy is similar to comprehension accuracy, albeit systematically slightly lower. Triphones and quadraphones again outperform biphones by a large margin. For the training data, triphones are somewhat less accurate than quadraphones.  Interestingly, in order to predict new forms in the validation data, triphones outperform quadraphones. Clearly, triphones offer better generalizability compared to quadraphones, suggesting that we are overfitting when modeling with quadraphones as cues.   Accuracy under the {\tt val\_newform} criterion is quite low, which is perhaps not unexpected given the uncertainty that characterizes native speakers' intuitions about the forms of novel words \citep{mccurdy2020inflecting}. In Section~\ref{sec:wugs} we return to this low accuracy, and consider in further detail generated novel forms and the best supported top candidates.

\subsubsection{Syllable-based representations}\label{sec:syllable}

Instead of using n-phones, the unit of analysis can be a combination of $n$ syllables.  The motivation for using syllables is that some suprasegmental features, such as lexical stress in German, are bound to syllables. Although stress information is not considered in the current simulation experiments,  suprasegmental cues can incorporated \citep[see][for an implementation]{Chuang:Bell:Banke:Baayen:2020}.

As for n-phones, when using n-syllables, we have to choose a value for the unit size $n$. For the word {\em Aale}, the bi-syllable cues are {\tt \#-a}, {\tt a-l@}, and {\tt l@-\#}, with ``-" indicating syllable boundary.  When unit size equals two, there are in total 8,401 unique bi-syllable cues. For tri-syllables, the total number of unique cues increases to 10,482. Above, we observed that the model was already overfitting with 9,068 unique quadraphone cues. We therefore do not consider tri-syllable cues, and only present modeling results for bi-syllable cues\footnote{Even though the number of bi-syllables is close to that of quadraphones, the fact that quadraphones still outnumber bi-syllables suggests that quadraphones have captured within-syllable phone collocations that are not available in bi-syllable cues. These further fine-grained cues might include, for example, consonant clusters, as in \underline{\it Spr}ache `language'.}.  

As shown in the fourth row of Table \ref{tab:form_res}, comprehension accuracy (for bi-syllables) for the training data is almost error-free, 99\%, the highest among all the cue representations. For the validation data, the overall accuracy is also high, 93\%. This is again due to the high accuracy for the seen forms ({\tt val\_lenient} = 99\%). Only one fifth of the unseen forms, however, is recognized successfully ({\tt val\_newform} = 20\%).  Production  accuracies for the training and validation data are 95\% and 63\% respectively. The model again performs well for homophones ({\tt val\_lenient} = 69\%) but fails to produce unseen forms ({\tt val\_newform} = 0.3\%).  This extremely low accuracy is in part due to the large number of cues that appear only in the validation dataset (325 for bisyllables, but only 23 for triphones).  Since such novel cues do not receive any training, words with such cues are less likely to be produced correctly. We will come back to the issue of novel cues in Section \ref{sec:careful_split}.  For now, we conclude that triphone-based form vectors are a good choice as they show a good balance of comprehension and production accuracy on training and validation data.

\subsection{Semantic representation}\label{sec:semantics}

There are many ways in which words' meanings can be represented numerically. The simplest method is to use one-hot encoding (i.e. a binary vector where a single value/bit is set to one), as implemented in the naive discriminative learning model proposed by \citep{Baayen:Milin:Filipovic:Hendrix:Marelli:2011}. One-hot encoding, however, misses out on the semantic similarities between lemmas:  all lemmas receive meaning representations that are orthogonal.
Instead of using one-hot encoding, semantic vectors can also be derived by turning words' taxonomies in WordNet into binary vectors with multiple bits on \citep[details in][]{Chuang:Bell:Banke:Baayen:2020}. In what follows, however, we work with  real-valued semantic vectors,  known as `word embeddings' in natural language processing.  
Semantic vectors can either be simulated, or derived from corpora using methods from distributional semantics \citep[see, e.g.][]{Landauer:Dumais:1997,mikolov2013distributed}.

\subsubsection{Simulated semantic vectors}\label{sec:simulated_vec}

When corpus-based semantic vectors are unavailable, semantic vectors can be simulated. The {\bf JudiLing} package enables the user to simulate such vectors using normally distributed random numbers for content lexemes and for inflectional functions.  By default, the dimension of the semantic vectors is set to be identical to that of the form vectors. 

The semantic vector for an inflected word is obtained by summing the vector of its lexeme and the vectors of all the pertinent inflectional functions. As a consequence, all vectors sharing a certain inflectional feature are shifted in the same direction in semantic space. By way of example, consider the German plural dative of \textit{Aal} `eel', \textit{Aalen}. We compute its semantic vector by adding the semantic vector for \textsc{plural} and \textsc{dative} to the lemma vector $\overrightarrow{Aal}_{lemma}$: 
\begin{align*}
   \overrightarrow{Aalen}_{dat.pl} = \overrightarrow{Aal}_{lemma} + \overrightarrow{\textsc{plural}} + \overrightarrow{\textsc{dative}}
\end{align*}
The corresponding singular dative \textit{Aal} can be coded as:
\begin{align*}
    \overrightarrow{Aal}_{dat.sg.} = \overrightarrow{Aal}_{lemma} + \overrightarrow{\textsc{singular}} + \overrightarrow{\textsc{dative}}
\end{align*}
Alternatively, the singular form could be coded as unmarked, following a privative opposition approach:
\begin{align*}
    \overrightarrow{Aal}_{dat.sg.} = \overrightarrow{Aal}_{lemma} +  \overrightarrow{\textsc{dative}}
\end{align*}
For the remainder of the paper, we treat number as an equipollent opposition. Finally, a small amount of random noise is added to each semantic vector (\textit{M} 0, \textit{SD} 1; compare this to \textit{M} 0, \textit{SD} 4 for lexeme and inflectional vectors), as an approximation of further semantic differences in word use other than number and case (see \citet[e.g. p.44ff.]{sinclair1991corpus}\footnote{Our approach of adding small semantic differences to individual word forms does probably not do justice to \citet{sinclair1991corpus}'s view that word forms can have completely idiosyncratic meanings, since we still assume commonalities across word forms such as e.g. a shared meaning of plurality. We hope to be able to address this issue in future research.}, \citet{tognini2001corpus} and further discussion below). The results reported thus far were all obtained with simulated vectors. 

It is worth noting that when working with simulated semantic vectors, the meanings of lexemes will still be orthogonal, and that as a consequence, all similarities between semantic vectors originate exclusively from the semantic structure that comes from the inflectional system.

\subsubsection{Empirical semantic vectors}\label{sec:emp_vec}

A second possibility for obtaining semantic vectors is to derive them from corpora.  \citet{Baayen:Chuang:Shafei:Blevins:2019} constructed semantic vectors from the TASA corpus \citep{ivens1991tasa}, in such a way that semantic vectors were obtained not only for lexemes but also for inflectional functions.  With their semantic vectors, the semantic vector of {\em Aalen} can be straightforwardly constructed from the semantic vectors of {\em Aal}, {\sc plural}, and {\sc dative}. 

However, semantic vectors that are created with standard methods from machine learning, such as \textit{word2vec} \citep{mikolov2013efficient}, \textit{fasttext} \citep{bojanowski2017enriching} or \textit{GloVe} \citep{pennington2014glove}, can also be used (albeit without semantic vectors for inflectional features; see below).  In what follows, we illustrate this for 300-dimensional vectors generated with \textit{word2vec}, trained on the German Wikipedia \citep{yamada2020wikipedia2vec}.  For representing words' forms, we used triphones. 

The model in general performs well for the training data (Table \ref{tab:form_res}). For the validation data, while the homophones are easy to recognize and produce, the unseen forms are again prohibitively difficult. Interestingly, if we compare the current results with the results of simulated vectors (cf. second row, Table \ref{tab:form_res}), we observe that while the {\tt train} and {\tt val\_all} accuracies are fairly comparable for the two vector types, their {\tt val\_newform} accuracies nonetheless differ. Specifically, understanding new forms is substantially more accurate with simulated vectors (51\% {\em vs.} 0.3\%), whereas {\em word2vec} embeddings yield slightly better results for producing new forms (21\% {\em vs.} 25\%). 

To understand why these differences arise, we note, first, that lexemes are more similar to each other than is the case for simulated vectors (in which case lexemes are orthogonal), and second, that \textit{word2vec} semantic vectors are exactly the same for each set of homophones within a paradigm, so that inflectional structure is much less precisely represented.  This lack of inflectional structure may underlie the inability of the model to understand novel inflected forms correctly.  Furthermore, the lack of differentiation between homophones simplifies the mapping from meaning to form, leading to more support from the semantics for the relevant triphones, which in turn facilitates synthesis by analysis.

In addition, we took the {\em word2vec} vectors, and reconstructed from these vectors the vectors of the lexemes and of the inflectional functions.  For a given lexeme, we created its lexeme vector by averaging over the vectors of its inflectional variants\footnote{Note that these vectors are not sense-disambiguated, so that the they can cover homophonous forms from various paradigm cells.}.  For plurality, we averaged over all vectors of forms that can be plural forms.  Using these new vectors, we constructed semantic vectors for a given paradigm cell by adding the semantic vector of the lexeme and the semantic vectors for its number and case values. The mean correlation between the new ``analytical'' {\em word2vec} vectors and the original empirical vectors was 0.79 ($sd = 0.076$).  Apparently, there is considerable variability in how German inflected words are actually used in texts, a finding that has also emerged from corpus linguistics \citep{sinclair1991corpus,tognini2001corpus}.  The idiosyncracies in the use of individual inflected forms renders the comprehension of an unseen, but nevertheless also idiosyncratic, inflected word form extremely difficult.  From this we conclude that the small amount of noise that we added to the simulated semantic vectors is likely to be unrealistically small compared to real language use. 

Interestingly, semantic similarity facilitates the production of unseen forms. A Linear Discriminant Analysis (LDA) predicting nine plural classes (the eight sub-classes presented in Table \ref{tab:plural_classes} plus one `other' class) from the {\em word2vec} semantic vectors has a prediction accuracy of 62.7\% (50.5\% under leave-one-out cross validation). Conducting 10-fold cross-validation with Support Vector Machine (SVM) yields an average accuracy of 56.7\%, considerably higher than the percentage of the majority choice (the -n plural class, 35.6\%). Apparently,  semantically similar words tend to inflect similarly. When a novel meaning is encountered in the validation set, it is therefore possible to predict to some extent its general form class.  Given the similarities between LDA and regression, the same kind of information is likely captured by LDL. 

\subsection{Missing forms and missing semantics}\label{sec:missing}

Evaluation on held-out data is a means for assessing the productivity of the network. However, it often happens during testing that the model is confronted with novel, unseen cues, or with novel, unseen semantics. Here, linguistically and cognitively motivated choices are required. 

\subsubsection{Novel cues} \label{sec:careful_split}

For the cross-validation results presented thus far, the validation data comprise a random selection of words. As a consequence, there often are novel cues in the validation data that the model has never encountered during training. The presence of such novel cues is especially harmful for production. As mentioned in Section \ref{sec:syllable}, the model with bi-syllables as cues fails to produce unseen forms, due to the large number of novel cues in the validation data.  

What is the theoretical status of novel cues?  To answer this question, first consider that actual speakers  rarely encounter new phones or new phone combinations in their native languages. Furthermore, novel sounds encountered in loan words are typically assimilated into the speaker's native phonology.\footnote{Note that such assimilation effects could be modeled using real acoustic input (i.e. audio files) with LDL-AURIS \citep{ShafaeiBajestan:2021}. Here, unseen sounds would presumably be assimilated to the closest seen sounds, similar to human performance. Of course, given sufficient training data, such a model would over time also be able to acquire the new sounds. We have, however, restricted ourselves to modeling using letter/phone representations.}   Also, many cues that are novel for the model actually occur not only in the held-out nouns, but also in verbs, adjectives, and compounds that the model has no experience with.  Thus, the presence of novel cues is in part a consequence of modeling only part of the German lexicon.  

Since novel cues have zero weights on their efferrent connections (or, equivalently, zero beta coefficients), they are completely inert for prediction. One way to address this issue is to select the held-out data with care.  Instead of randomly holding out words, we make sure that in the validation data all cues are already present in the training data.   We therefore split the dataset into 80\% training and 20\% validation data, but now making sure that there are no novel triphone cues in the validation dataset. Among the 3,629 validation words, 3,331  are homophones, and 298 are unseen forms.  We note that changing the kind of cues used typically has consequences for how many datapoints are available for validation.  When bi-syllables are used instead of triphones,  due to the sparsity of bi-syllable cues, we have to increase the percentage of validation data to include sufficient numbers of unseen forms. Even for 65\% training data and 35\% validation data,  the majority of validation data are homophones (98.5\%), and only 76 cases represent unseen forms (with only known cues). 

\begin{table}[tb!]
{\small
    \centering
    \begin{tabular}{l|c|c|cc|c|c|cc} \hline
                   & \multicolumn{4}{c}{comprehension} & \multicolumn{4}{|c}{production} \\ \cline{2-9}
                   & train & val\_all & val\_lenient & val\_newform & train & val\_all & val\_lenient & val\_newform  \\ \hline
       triphone    &  93\% & 88\% & 91\% & 52\% & 85\% & 63\% & 67\% & 17\% \\
       bisyllable  &  99\% & 99\% & 99\% & 61\% & 95\% & 52\% & 52\% & 12\%  \\
       \hline
    \end{tabular}
}
    \caption{Comprehension and production accuracy for train and validation datasets, which are split in such a way that no novel cues are present in the validation set. Both the triphone and bisyllable models make use of simulated semantic vectors.}
    \label{tab:novel_res}
\end{table}

For the triphone model (top row, Table \ref{tab:novel_res}), for both comprehension and production, the {\tt train}, {\tt val\_all} and {\tt val\_lenient} accuracies are similar to the results presented previously (Table \ref{tab:form_res}). For the evaluation of unseen forms ({\tt val\_newform}), there is only a slight improvement for comprehension (from 51\% to 52\%); for other datasets, the improvement can be larger.  However, for production, {\tt val\_newform} becomes worse (decreasing from 21\% to 17\%).  The reason is that even though all triphone cues of the validation words are present in the training data, they obtain insufficient support from the semantics.  The solution here is to allow a small number of triphone cues with weak support (below the threshold $\theta$) to be taken into account by the algorithm that orders triphones into words. This requires turning on the {\tt tolerance} mode in the {\tt learn\_paths} function of the {\bf JudiLing} package. By allowing at most two weakly supported triphones to be taken into account, production accuracy for unseen forms increases to 57\%.

The bi-syllable model benefits more from the removal of novel cues in the validation data. Especially for comprehension, the accuracy of unseen forms reaches  61\%, compared to 20\% with random selection. For production, we observe a non-negligible improvement as well (from 0.3\% to 12\%). Further improvements are expected when tolerance mode is used, but given the large number of bi-syllables, this comes at considerable computation costs.  In other words, bi-syllables provide a model that is an excellent memory, but a memory with very limited productivity specifically for production.

\subsubsection{Unseen semantics}\label{sec:wugs}

In real language, speakers seldom encounter words that are completely devoid of meaning: even novel words are typically encountered in contexts which narrow down their interpretation. In the wug task, by contrast, participants are often confronted with novel words presented without any indication of their meaning, as, for instance, in the experiment on German nouns reported by \citet{mccurdy2020inflecting}.  Within the framework of the discriminative lexicon, this raises the question of how to model the wug task, as the model has no way to produce inflected variants without semantics.

For modeling the wug task, and comparing model performance with that of German native speakers, we begin with observing that the comprehension system generates meanings for nonwords. \citet{chuang2020processing} showed that measures derived from the semantic vectors of nonwords were predictive for both reaction times in an auditory lexical decision task and for nonwords' acoustic durations in a reading task. In order to model the wug task, we therefore proceeded as follows: 
\begin{enumerate}
\item We first simulated a speaker's lexical knowledge prior to the experiment by training a comprehension matrix using all the words described in Section~\ref{sec:modelling}.  Here,  we made use of simulated semantic vectors.
\item We then used the resulting comprehension network to obtain semantic vectors $\bm{s}_\text{nom.sg}$ for the nominative singular forms of the nonwords by mapping their cue vectors into the semantic space, resulting in semantic vectors $\bm{s}_\text{nom.sg}$.
\item Next, we created the production mapping from meaning to form, using not only all real words but also the nonwords (known only in their nominative singular form).  
\item Then, we created the semantic vectors for the plurals ($\bm{s}_\text{nom.pl}$) of the nonwords by adding the plural vector to their nominative singular vectors while subtracting the singular vector.
\item Finally, these plural semantic vectors were mapped onto form vectors ($\hat{\bm{c}}_\text{nom.pl}$) using the production matrix, in combination with the \texttt{learn\_paths} algorithm that orders triphones for articulation.
\end{enumerate}

We applied these modeling steps to a subset of the experimental materials provided by \citet{marcus1995german} \citep[reused by][]{mccurdy2020inflecting},  in order to compare  model predictions  with the results of  \citet{mccurdy2020inflecting}.  The full materials of \citet{marcus1995german} contained nonwords  that were set up such that only half of them had an existing rhyme in German. We restricted ourselves to the nonwords with existing rhymes, first, because non-rhyme words have many cues that are not in the training data; and second, because, as noted by \citet{zaretsky2015no}, many of the non-rhyme words have unusual orthography and thus are strange even for German speakers.  Furthermore,  many of the non-rhyme nonwords share endings and therefore do not provide strong data for testing model predictions.

\citet{mccurdy2020inflecting} presented nonwords visually and asked participants to write down their plural form.
To make our simulation more comparable to their experiment, in the following we made use of letter trigrams rather than triphones. We represented words without their articles, as the wug task implemented by \citet{mccurdy2020inflecting} presented the plural article as a prompt for the plural form;  participants thus produced bare plural forms.  For assessing what forms are potential candidates for production, we examined the set of candidate forms, ranked by how well their internally projected meanings (obtained with the synthesis-by-analysis algorithm, see Section~\ref{sec:ldl}), correlated with the targeted meaning $\bm{s}_\text{nom.pl}$.  We then examined the best supported candidates as possible alternative plural forms.

The model provided a plausible plural form as the best candidate in 7 out of 12 cases. Five of these belonged to the \textit{-en} class. A further plausible candidate was also only provided in 5 of the cases. The lack of diversity as well as the bias for \textit{-en} plurals does not correspond to the responses given by German speakers in \citet{mccurdy2020inflecting}.

Upon closer inspection, it turns out that a more variegated wug performance can be obtained by changing two parameters.  First, we replaced letter trigrams by letter bigrams. This substantially reduces the number of n-grams that are present in the nonwords  but that do not occur in the training data.  Second, we made a small but important change to how semantic vectors were simulated.  The default parameter settings provided with the \textbf{JudiLing} package generate semantic vectors with the same standard deviation for both content words and inflectional features.  Therefore, the magnitudes of the values in semantic vectors is very similar for content words and inflectional features. Since words are inflected for case and number, their semantic vectors are numerically dominated by the inflectional vectors.  To enhance the importance of the lexemes, and to reduce the dominance of the inflectional functions, we reduced the standard deviation (by a factor of $\frac{1}{10}$) when generating the semantic vectors for number and case.  As a consequence, the mean of the absolute values in the plural vector decreased from 3.25 to 0.32.  (Technical details are provided in the supplementary materials.) With these two changes, the model generated a more diverse set of plural nonword candidates (Table~\ref{tab:model_outputs}). Model performance is now much closer to the performance of native speakers as reported by \citep{zaretsky2013acquisition, mccurdy2020inflecting}.  

\begin{table}[tb!]
    \centering
    \begin{tabular}{c|c|c|c|c|c}
    \hline
Bral & Kach & Klot & Mur & Nuhl & Pind \\
\hline
Bralen & Kachen & Klot & Muren & Nuhlen & Pinden\\
Bral & Kach & *Klotten & Murn & Nuhl & Pind \\
*Bralenen & Kacher & *Klotte & Mur & Nuhle & Pinder\\
*Bralern & Kache & *Klotter & *Murnen & *Nuhlern & Pinde\\
Braler & *Kachern & *Klieloten & Murer & *Nuhlere & *Pindern\\
\hline
\end{tabular}\vspace{0.1cm}
\begin{tabular}{c|c|c|c|c|c}
    \hline
Pisch & Pund & Raun & Spand & Spert & Vag\\
\hline
Pischen & Punden & Raunen & *Spanend & Sperten & Vag\\
Pisch & *Punend & Raun & Spand & Sperte & Vagen\\
Pischer & Pund & *Raunern & *Spanende & Sperter & Vage\\
Pische & Punde & Rauner & *Spanenden & *Spererten & Vager\\
*Pischern & *Pundene & Raune & *Spatend & *Spererte & *Vagern\\
\hline
\end{tabular}
    \caption{First five candidates for the plural forms of nonwords. Forms that are implausible as plurals are marked with an asterisk.}
    \label{tab:model_outputs}
\end{table}

The model also produces some implausible plural candidates, all of which however are phonotactically legal;  these are marked with an asterisk in Table~\ref{tab:model_outputs}.  Sometimes a plural marker is interfixed instead of suffixed (e.g., Spand, Span-\textit{en}-d; Pund, Pun-\textit{en}-d). Almost all words have a candidate which shows double plural marking (e.g. Bral, Bral-\textit{en}-\textit{en}; Nuhl, Nuhl-\textit{er}-\textit{e}; Pind, Pind-\textit{er}-\textit{n}; cf. Dutch kind-\textit{er}-\textit{en}), or a mixture of both (e.g. Span, Span-\textit{en}-d-\textit{e}; Spert, Sper-\textit{er}-t-\textit{en}). For Klot, doubling of the \textit{-t} can be observed, as this form is presumably more plausible in German (e.g. Motte (`moth'), Gott (`god'), Schrott (`scrap, rubbish')). One plural has been attracted to an existing singular (Spand, Spaten-d).  Apparently, by downgrading the strength (or more precisely, the L1-norm) of the semantic vectors of inflectional functions, the model moves in the direction of interfixation-like changes. 

The model does not produce a single plural form with an umlaut, even though in corpora umlauted plurals are relatively frequent \citep[see e.g.][]{gaeta2008deutsche}.  Interestingly, the German speakers in \citet{mccurdy2019german} also tended to avoid umlauted forms (with the exception of \textit{Kach} $\rightarrow$ \textit{Kächer}). Interestingly, children at the age of 5 also tend to avoid umlauts when producing plurals for German nonwords, but usage increases for 7-year-olds and adults \citep{vijver2014developing}.

Finally, most nonwords have a plural in {\em -en} as one of the candidates (10 out of 12 cases), with as runners-up the {\em -e} plural (8 out of 12 cases), and the {\em -er} plural (8 out of 12).  There is not a single instance of an {\em -s} plural, which fits well with the low prevalence (around 5\%) of {\em -s} plurals in the experiment of \citet{mccurdy2020inflecting}.

This simulation study shows that it is possible to make considerable headway with respect to modeling the wug task for German.  The model is not perfect, unsurprisingly, given that we have worked with simulated semantic vectors and estimates of nonwords' meanings.  It is intriguing that a strong weight imposed on the stem shifts model performance in the direction of interfixation-like morphology.  However, the model has no access to information about words' frequency of use, and hence is blind to an important  factor shaping human learning (see Section~\ref{sec:kind_of_learning} for further discussion). Nevertheless, the model does appear to mirror the uncertainties of German speakers fairly well. 

\subsection{Words in context}\label{sec:context}

Thus far, we have modeled words in isolation.  However, in German, case and number information is to a large extent carried by preceding determiners. In addition, in actual language use, a given grammatical case denotes one of a wide range of different possible semantic roles.  The simplifying assumption that an inflectional function can be represented by a single vector, which may be reasonable for grammatical number, is not at all justified for grammatical case.  In this section, we therefore explore how context can be taken into account.   We first present  modeling results of nouns learned together with their articles. Next, we break down grammatical cases into actual semantic functions, and show how we can begin to model the noun declension system with more informed semantic representations.

\subsubsection{Articles}\label{sec:def_article} 

We first consider definite articles. Depending on gender and case, a noun can follow one of the six definite articles in German --- der, die, das, dem, den, des.  We added these articles, transcribed in DISC notation,  before the nouns. Although in writing articles and nouns are separated by a space character (e.g. \texttt{der Aal}), to model auditory comprehension we removed the space character (e.g., \texttt{deral}).  By adding the articles to the noun forms,  the number of homophones in our dataset was reduced to a substantial extent, whereas the number of unique word forms more than doubled (from 5,427 to 12,798). 

In the first set of simulations we used the same semantic vectors as we did previously for modeling isolated words. That is, the meanings of the definite articles are not taken into account in the semantic vectors, as all forms would be shifted in semantic space in the exactly the same way.  After including articles, the validation data now only contained 3,982 homophones, but the number of unseen forms increased to 3,260. Using triphones as cues, we ran two models, one with simulated vectors and the other with {\em word2vec} semantic vectors.  For simulated vectors the results (Table \ref{tab:article_res}) are generally similar to those obtained without articles (Table \ref{tab:form_res}). However, if we look at the evaluation of comprehension with the strict criterion (according to which recognizing a homophone is considered incorrect), without articles {\tt val\_strict} is 6\%, whereas it is 34\% with articles.   The generalizability of the model also improves as the number of homophones in the dataset decreases.  Even though there are more unseen forms in the current dataset with articles than in the original one without articles, the {\tt val\_newform} for comprehension increases by 12\% from 51\% to 63\%.

\begin{table}[]
{\small
    \centering
    \begin{tabular}{l|c|c|cc|c|c|cc} \hline
                   & \multicolumn{4}{c}{comprehension} & \multicolumn{4}{|c}{production} \\ \cline{2-9}
                   & train & val\_all & val\_lenient & val\_newform & train & val\_all & val\_lenient & val\_newform  \\ \hline
       simulated    &  94\% & 76\% & 92\% & 63\% & 81\% & 37\% & 57\% & 19\% \\
       {\em word2vec}  &  91\% & 69\% & 81\% & 58\% & 48\% & 14\% & 28\% & 1\%  \\ \hline
       def + indef &  94\% & 80\% & 93\% & 64\% & 82\% & 40\% & 60\% & 15\% \\
       \hline
    \end{tabular}
}
    \caption{Comprehension and production accuracy for train and validation datasets with articles. All three simulations use triphones as cues. The first two rows present results with simulated vectors and {\em word2vec} embeddings as semantic representations. The simulation presented in the bottom row also makes use of simulated vectors, but includes both definite and indefinite articles. }
    \label{tab:article_res}
\end{table}

When using {\em word2vec} embeddings, adding articles to form representations also improved the comprehension of unseen forms: the {\tt val\_newform} astonishingly increased from 0.3\% to 58\%.  Without articles, homophones all shared the same form representations and exactly the same {\em word2vec} vectors.  As a consequence, many triphone cues were  superfluous and not well-positioned to discriminate between lemma or inflectional meanings.  Now, with the addition of articles, the form space is better discriminated. With an increased number of triphone cues, the model is now able to predict and generalize more accurately for comprehension. However, for production, model performance is generally worse when articles have to be produced.  For the training data, for instance, production accuracy drops from 97\% (without articles) to 48\%. This is of course unsurprising. In the simulation with articles, the semantic representations remain the same, but now identical semantic vectors have to predict more variegated triphone vectors. The learning task has become more challenging, and inevitably resulted in less accurate performance. Replacing the contextually unaware {\em word2vec} vectors by contextually aware vectors obtained using language models such as BERT \citep{devlin2018bert, miaschi2020contextual} should alleviate this problem.

We can test the model on more challenging data by including indefinite articles (ein, eine, einem, einen, einer, eines), and creating two additional semantic vectors, one for definiteness and one for indefiniteness. This doubles the size of our dataset:  half of the words are preceded by definite articles, and the other half  by indefinite articles. However, because German indefinite articles are restricted to singular forms, only indefinite singular forms are preceded by indefinite articles.  On the meaning side, the $\overrightarrow{\textsc{definite}}$ vector is added to the semantic vectors of words preceded by definite articles, and the $\overrightarrow{\textsc{indefinite}}$ vector is added to vectors for words preceded by either indefinite articles in the singular, or no article in the plural.

The validation data of this dataset confronts the model with in total 3,982 homophones and 3,260 unseen forms. Homophones comprise slightly more words with indefinite articles (57\%) whereas unseen forms comprise slightly more definite articles (59\%). The results, presented in the bottom row of Table \ref{tab:article_res}, are very similar to those with only definite articles (top row). Closer inspection of the results for the validation data shows that for comprehension, accuracies do not differ much across definite and indefinite forms. For production, however, especially for unseen forms, the accuracy for definite articles is twice higher than that for indefinite articles (20\% and 9\%, averaging out to 15\%). This is a straightforward consequence of the much more diverse realizations of indefinite nouns.  For definite nouns, the possible triphone cues at the first two positions in the word are always limited to the triphone cues of the six definite articles. For indefiniteness, however, in addition to the six indefinite articles, initial triphone cues also originate from words' stems --- indefinite plural forms are realized without articles.  The mappings for production are thus faced with a more complex task for indefinites,  and the model is therefore more likely to fail on indefinite forms.

\subsubsection{Semantic roles} \label{sec:semantic_roles}

The simulation studies thus far suggest it is not straightforward to correctly comprehend a novel German word form in isolation, even when articles are provided. This is perhaps not that surprising, as in natural language use, inflected words appear in context, and usually realize not some abstract case ending, but a specific  semantic role \citep[also called \textit{thematic role}, see, e.g.,][]{harley2010thematic}.  For example, a word in the nominative singular might express a theme, as \textit{der Apfel} in \textit{Der Apfel fällt vom Baum.} (`The apple falls from the tree'), or it might express an agent as \textit{der Junge} in \textit{Der Junge isst den Apfel.} (`The boy eats the apple.'). Exactly the same lemma,  used with exactly the same case and number, may still realize very different semantic roles. Consider the two sentences \textit{Ich bin bei der Freundin} (`I'm at the friend's') and \textit{Ich gebe der Freundin das Buch.} (`I give the book to the friend`). \textit{der Freundin} is dative singular in both cases, but in the first sentence, it expresses a location while in the second it represents the beneficiary or receiver. Semantic roles can also be reflected in a word's form, independently of case markers. For example, German nouns ending in \textit{-er} are so-called ``Nomina Agentis'' \citep{baeskow2011abgeleitete}. As pointed out by \citet{Blevins:2016}, case endings are no more (or less) than markers for the intersection of form variation and a distribution class of semantic roles.  Since within the framework of the DLM, the aim is to provide mappings between form and meaning, a case label is not a proper representation of a word's actual meaning.  All it does is specify a range of meanings that the form can have, depending on context.  Therefore, even though we can get the mechanics of the model to work with case specifications, doing so clashes with the `discriminative modeling' approach.  In what follows, we therefore implement mappings with somewhat more realistic semantic representations of German inflected nouns.

Our starting point is that in German, different cases can realize a wide range of semantic roles.  For our simulations, we restrict ourselves to some of the most prominent semantic roles for each case (Table \ref{tab:sem_roles}). Even though these clearly do not reflect the full richness of the semantics of German cases, they suffice for a proof-of-concept simulation.

\begin{table}[tb!]
    \centering
    \begin{tabular}{l|r} \hline
        Case & Semantic roles \\
        \hline
        Nominative & agent (50\%), theme (40\%), patient (10\%) \\
        Genitive & possessive (90\%), partitive (10\%) \\
        Dative & beneficiary (50\%), location (50\%)\\
        Accusative & patient (40\%), motion (30\%), experiencer (30\%) \\ \hline
    \end{tabular}
    \caption{Probabilities of semantic roles by cases in the German noun system. Semantic roles are informed by \citet{Schulz1981}. Percentages are simulated and do not necessarily reflect corpus-frequencies of the respective semantic role.}
    \label{tab:sem_roles}
\end{table}

In order to obtain a dataset with variegated semantic roles, we expanded the previous dataset, with each word form (including its article) appearing with a specification of its semantic role, according to the probabilities presented in Table~\ref{tab:sem_roles}.  The resulting dataset had 45,605 entries, which we randomly split into 80\% training data and 20\% validation data. For generating the semantic matrix, we again used number, but instead of a case label, we provided the \textit{semantic role} as inflectional feature.   Comprehension accuracy on this data is comparable to the previous simulations: 89\% for the training data \texttt{train}, and 85\% \texttt{val\_lenient}. Comprehension accuracy on the validation set drops dramatically when we use strict evaluation (4\% accuracy). This is unsurprising given that it is impossible for the model to know which semantic role is intended when only being exposed to the word form and its article in isolation, without syntactic context.  Production accuracy is likewise comparable to previous simulations with \texttt{train} at 78\% and \texttt{val\_lenient} at 61\% (\texttt{val\_newform} 25\%). This simple result clarifies that in order to properly model German nouns, it is necessary to take the syntactic context in which a noun occurs into account.  Future research will also have to face the challenge of integrating words' individual usage profiles into the model (see also Section \ref{sec:simulated_vec} above).

\subsection{Incremental learning versus the end-state of learning} \label{sec:kind_of_learning}

In the simulation studies presented thus far, we made use of the regression method to estimate the mappings between form and meaning.  The regression method is strictly type based: the data on which a model is trained and evaluated consists of all unique combinations of form vectors $\bm{c}$ and semantic vectors $\bm{s}$. In this respect, the regression method is very similar to models such as AML, MBL, MGL, and to statistical analyses with the GLM or recursive partioning methods. However, word types (understood as unique sets $\{\bm{c}, \bm{s}\}$) are not uniformly distributed in language, and there is ample evidence that the frequencies with which word types occur co-determines lexical processing \citep[see, e.g.,][]{Baayen:Dijkstra:Schreuder:1997,Baayen:Milin:Ramscar:2016,Baayen:Wurm:Aycock:2007,tomaschek2018practice}.  While some formal theorists flatly deny that word frequency effects exist for inflected words \citep{Yang:2016}, others have argued that there is no problem with integrating frequency of use into formal theories of the lexicon \citep{Jackendoff:75,jackendoff2019texture}, and yet others have argued that it is absolutely essential to incorporate frequency into any meaningful account of language in action \citep{Langacker:1987,Bybee:2010}.

Within the present approach, effects of frequency of occurrence can be incorporated seamlessly by using incremental learning instead of the end-state of learning as defined by the regression equations \citep[see][for the convergence over learning time of incremental learing to the regression end-state of learning]{Danks2003,Evert:Arppe:2015,ShafaeiBajestan:2021}.  We illustrate this for our German nouns dataset with number and semantic role as crucial constructors of simulated semantic vectors. 

We begin with noting that word forms usually do not instantiate all possible semantic roles equally frequently.  For instance, a word such as \textit{der Doktor} (`doctor') will presumably occur mostly as \textit{agent} in the nominative singular form, rather than as \textit{theme} or \textit{patient}. If the model is informed about the probability distributions of semantic roles in actual language use (both in language generally, and lexeme-specific), it may be expected to make more informed decisions when coming across new forms, for instance, by opting for the best match given its past experience. 

Incremental learning with the learning rule of Widrow-Hoff makes it possible to start approximating human word-to-word learning as a function of experience.  As a consequence, the more frequent a word type occurs in language use, the better it can be learned: practice makes perfect.  This sets the following simulation study apart from models such as proposed by \citet{belth2021greedy} or \citet{mccurdy2020inflecting}, who base their training regimes strictly on types rather than tokens. 

In the absence of empirical frequencies with which combinations of semantic roles and German nouns co-occur, we simulated frequencies of use.\footnote{Though there are several semantic role labelers available for English (e.g. arising from the CoNLL-2004 and 2005 Shared Tasks (\url{https://www.cs.upc.edu/~srlconll/home.html})), there are --- to our knowledge --- currently no suitable taggers for German.} To do so, we proceeded as follows. First, we collected token frequencies for all our word forms from CELEX. Next, we assigned an equal part $\textit{freq}_p$ of this frequency count to each case/number cell realising this word form.  Third, for each paradigm cell, we randomly set to zero some semantic roles,  drawing from a binomial distribution with $n=1$, $p=\frac{1}{K}$, with $K$ the number of semantic roles for the paradigm cell (see Table~\ref{tab:sem_roles}).  In this way, on average,  one semantic role was omitted per paradigm cell. Finally, given a proportional frequency count $\textit{freq}_p$, the semantic roles associated with a paradigm cell received frequencies proportional to the percentages given in Table \ref{tab:sem_roles}.  Further details on this procedure are available in the supplementary materials,  a full example can be found in Table~\ref{tab:semantic_roles_freqs}.

\begin{table}[tb!]
    \centering
    \begin{tabularx}{\textwidth}{l|l|l|l|l|X|X} \hline
        word form & lemma & case & number & semantic role & form frequency & form+role\linebreak frequency  \\ \hline
        \hline
        Adresse & Adresse & nominative & singular & agent & 137 & 20\\
        Adresse & Adresse & nominative & singular & theme & 137 & 16\\
        Adresse & Adresse & nominative & singular & patient & 137 & 0\\
        Adresse & Adresse & genitive & singular & possessive & 137 & 0\\
        Adresse & Adresse & genitive & singular & partitive & 137 & 35\\
        Adresse & Adresse & dative & singular & beneficiary & 137 & 18\\
        Adresse & Adresse & dative & singular & location & 137 & 18\\
        Adresse & Adresse & accusative & singular & patient & 137 & 0\\
        Adresse & Adresse & accusative & singular & motion & 137 & 0\\
        Adresse & Adresse & accusative & singular & experiencer & 137 & 35\\ \hline
    \end{tabularx}
    \caption{Example of simulated frequencies for combinations of case and semantic role for the word form  ``Adresse''.}
    \label{tab:semantic_roles_freqs}
\end{table}

Having obtained simulated frequencies, we proceeded by randomly selecting 274 different lemmas (1,274 distinct word forms with definite articles included), in order to keep the size of the simulation down --- simulating with the Widrow-Hoff rule is computationally expensive. The total number of tokens in this study was 4,470. For the form vectors, we used triphones.   The dimension of the simulated semantic vectors was identical to that of the cue vectors. As before, the data was split into 80\% training and 20\% validation data. We followed the same procedure as in the previous experiments, but instead of computing the mapping matrices in their closed form (i.e. end-state) solution, we used incremental learning. 

While for comprehension, the implementation of the learning algorithm is relatively straightforward, this is not the case for production. The \texttt{learn\_paths} algorithm calculates the support for each of the n-grams,  for each possible position in a word. In the current implementation of {\bf JudiLing}, the calculation of positional support is not implemented for incremental learning. Therefore, we do not consider incremental learning of production here. 

Comprehension accuracy was similar to that observed for previous experiments.  Training accuracy when taking into account homophones was 85\%, validation accuracy on the full data was 79\% (\texttt{val\_lenient}).  Without considering homophones, validation accuracy drops substantially (\texttt{val\_strict} 7\%). This is unsurprising given that from the form alone it is impossible to predict the proper semantic role. 

\begin{figure}[tb!]
     \centering
     \includegraphics[width=\textwidth]{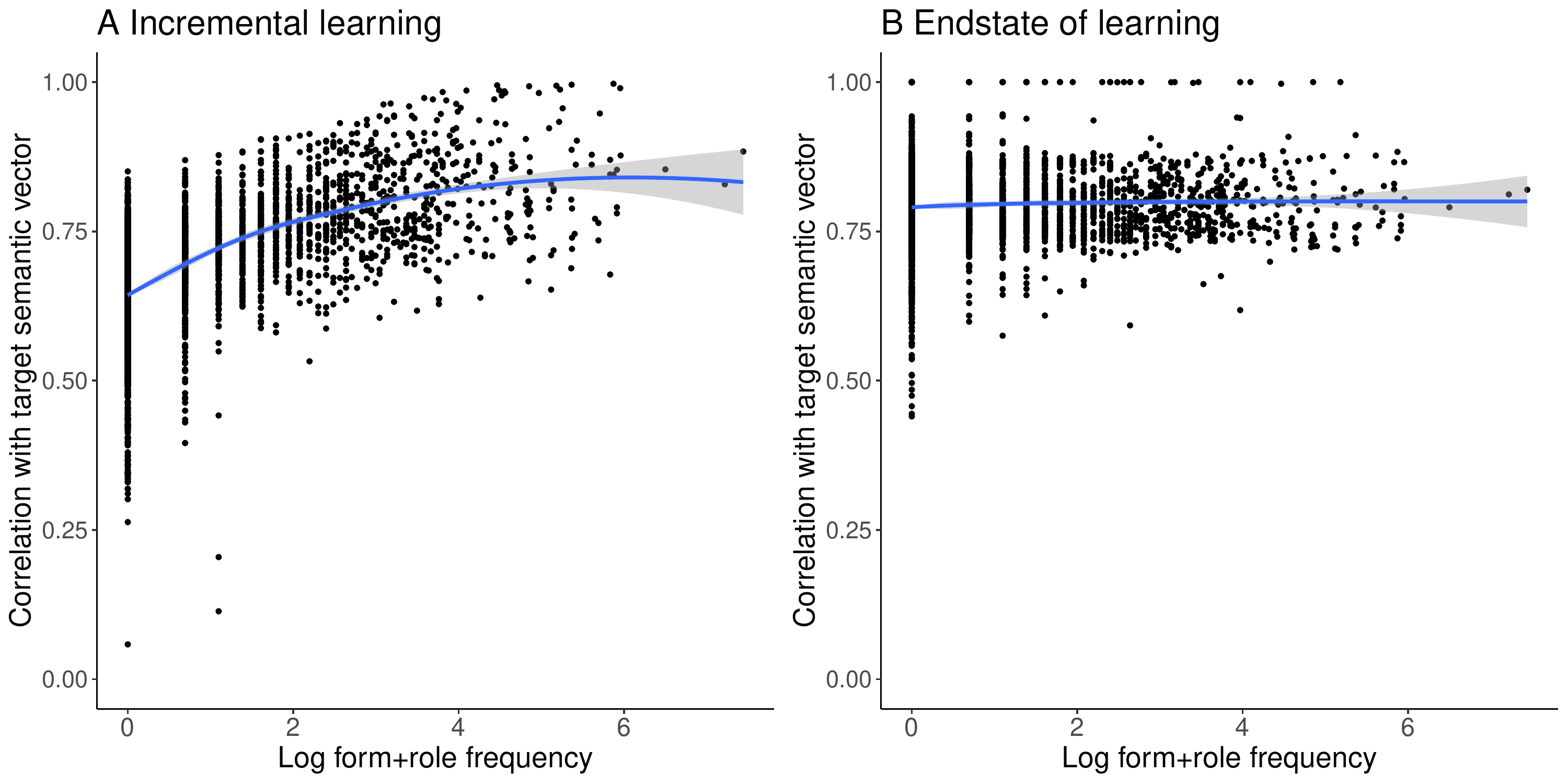}
     \caption{Correlation between the simulated frequency and correlation of the predicted semantic vector with its target. Generally, the more frequent a word form is,  the more accurate its semantic vector is predicted. The blue line indicates a loess smooth with a .95 confidence interval.} 
     \label{fig:freq_s_corr}
\end{figure}

The accuracy of the model's predictions is also closely linked to the frequencies with which words' form+role combinations are encountered in the training data.  If a word's form+role combination is very frequent, it is learned better.  Figure~\ref{fig:freq_s_corr} presents the correlations of words' predicted and targeted semantic vectors against their frequency of occurrence. The left panel presents the results for the incrementally learned model, the right panel for the end-state of learning. Clearly, after incremental learning the model predicts the semantics of more frequent form+role combinations more accurately. For the end-state of learning on the other hand, no such effect can be observed. These results clearly illustrate the difference between a token-based model and a typed-based model. 

\begin{figure}[tb!]
     
         \centering
         \includegraphics[width=\textwidth]{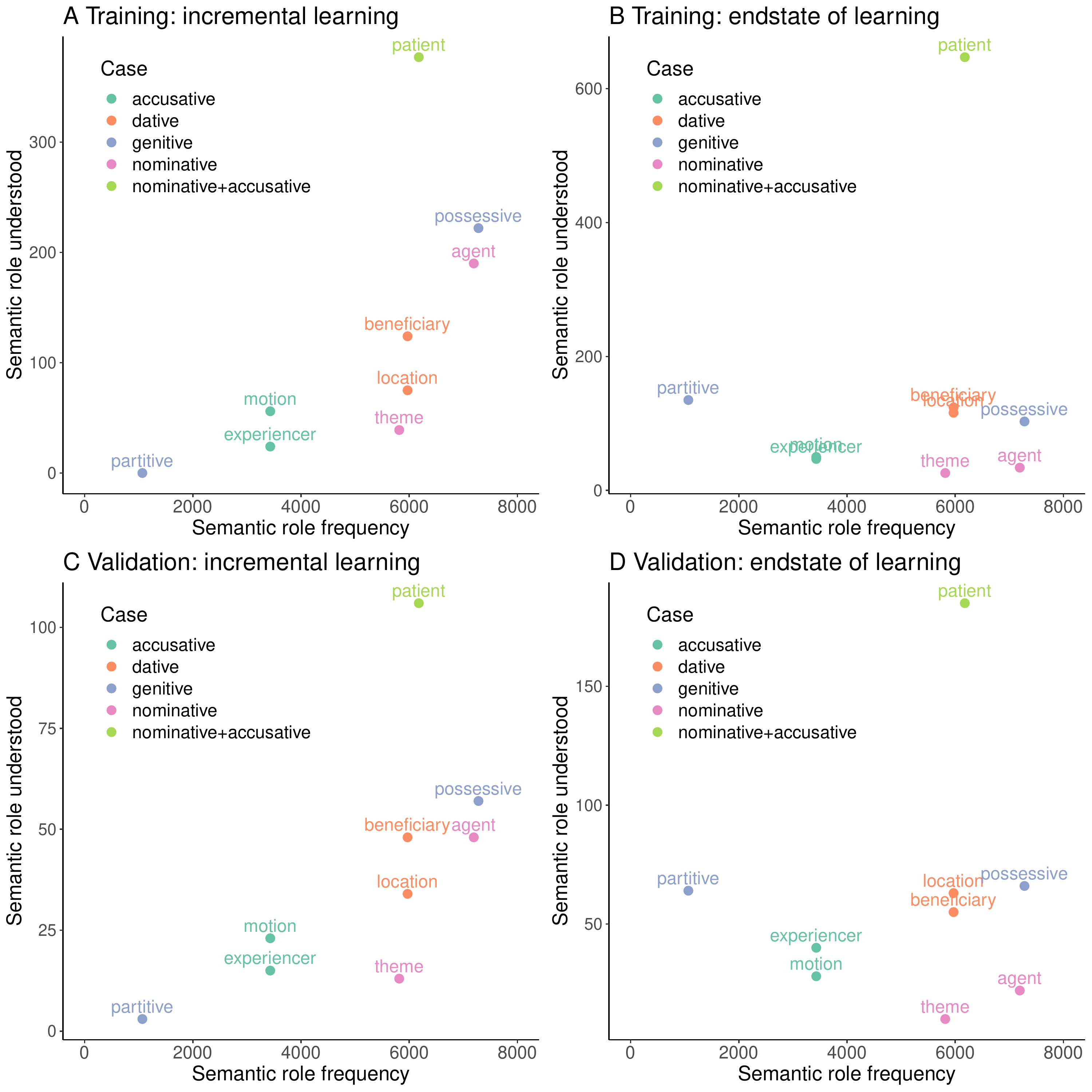}
        \caption{Counts of overgeneralization errors of semantic roles for training (top) and test data (bottom), for incremental learning (left) and the end-state of learning (right), conditional on the model having understood lexeme, number, and case correctly. 
        }
        \label{fig:errors}
\end{figure}

The effect of frequency of use on the kind of errors made by the model is also revealing. We zoom in on those cases where the model was able to correctly identify the lemma and paradigm cell of the word form, but did not get the semantic role right. Figure~\ref{fig:errors} provides scatterplots graphing the number of times a semantic role was (incorrectly) understood against the frequency of the form's semantic role, cross-classified by training method (incremental, left panels; end-state of learning, right panels) and by evaluation set (top panels: training data, bottom panels: validation data). For incremental learning, there is a positive correlation between the number of times a semantic role was (incorrectly) identified and the frequency of the semantic role in the training data.   Note that the relation is not linear, but curvilinear. A linear relation would have implied that a fixed  proportion of word forms would be incorrectly recognized, across all semantic roles.  What we see, by contrast, is that greater exposure in language use has an increasingly detrimental effect on learning, with more probable semantic roles being over-identified. Importantly,  for the end-state of learning, this curvilinear effect of frequency on learning is absent, with the {\sc patient} role representing an atypical outlier.  This outlier status is due to the \textit{patient} semantic role being realized by two cases: \textit{nominative} and \textit{accusative}.  As a consequence, it is not only frequent, but it is also predicted by many more different cues (especially cues from the articles) than is the case for other semantic roles. 

In other words, with incremental learning, strong frequency effects emerge, hand in hand with overgeneralization of semantic roles. (The study by \citet{Ramscar:Dye:McCauley:2013} makes the same point for irregular English noun plurals.) By contrast, for the end-state of learning, such effects are absent.  Mathematically, this makes sense: as experience (i.e., volume of training data) goes to infinity, all forms are learned an infinite number of times, and frequency is no longer distinctive.

\begin{figure}[tb!]
    \centering
    \includegraphics[width=.45\textwidth]{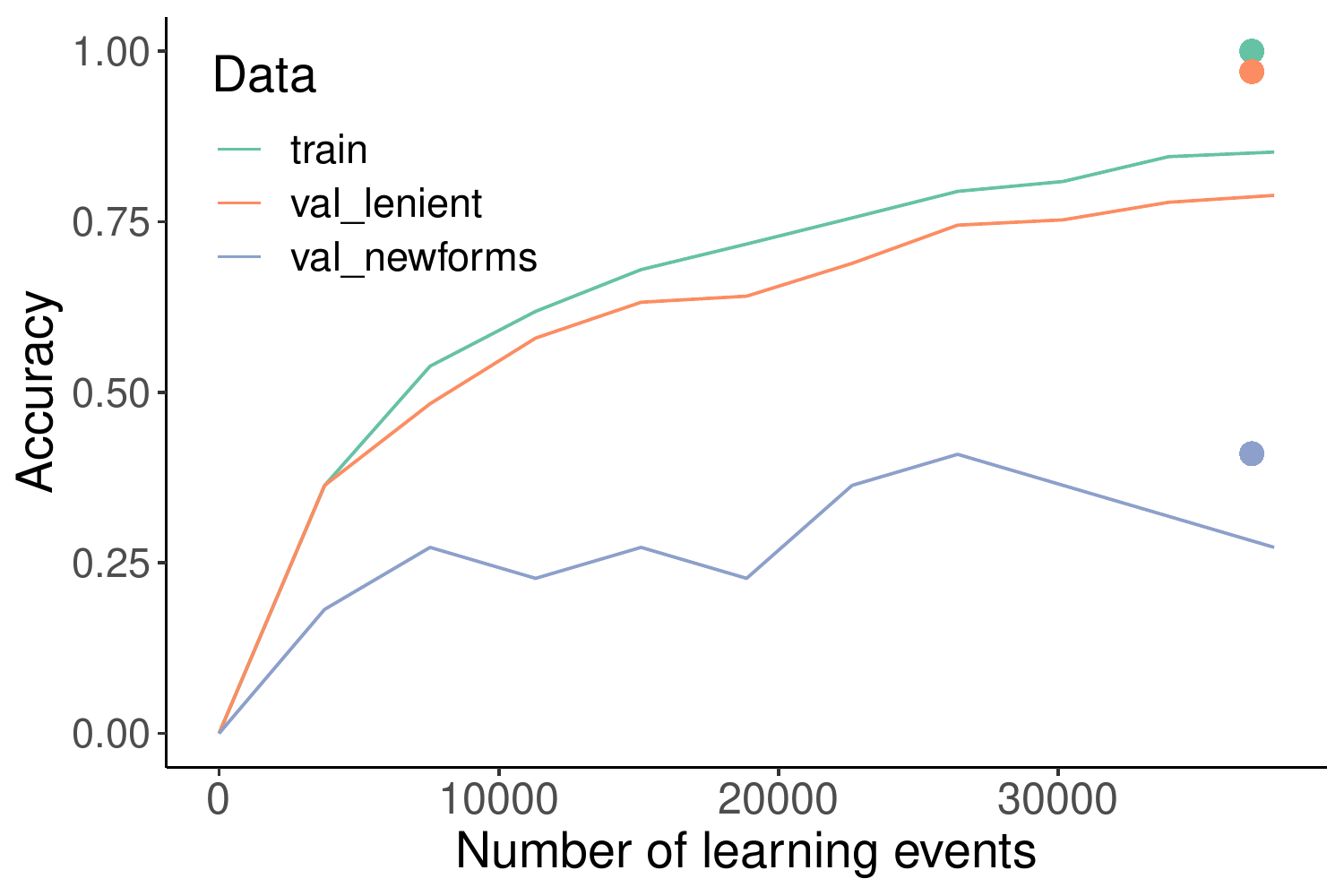}
    \caption{Comprehension accuracy over the course of learning. After a very fast increase in accuracy over the first 15,000 learning events, the amount of learning levels off. Points indicate the accuracy at the end-state of learning which the incremental model would reach eventually after an infinite number of learning events.
    }
    \label{fig:learning_trajectory}
\end{figure}

With incremental learning, it is also possible to follow the learning trajectory of the model.  Figure~\ref{fig:learning_trajectory} presents this trajectory at 10 evaluation points.  Learning proceeds rapidly during the first 15,000 learning events and slows down afterwards. Validation accuracy \texttt{val\_lenient} closely follows training accuracy, which is a straightforward consequence of the large numbers of homophones. \texttt{val\_newforms} on the other hand stays relatively low, in accordance with the semi-productivity of the German declension system.  

Note that in this simulation we only pass through the data once. If a word form has a form+role frequency of 1, it is only seen a single time during  training. As such, it is not possible for the model to reach accuracies as high as at the end-state of learning (indicated as dots in Figure~\ref{fig:learning_trajectory}), which would be reached eventually after an infinite number of passes through the data \citep{Danks2003,Evert:Arppe:2015,ShafaeiBajestan:2021}.  This sets our approach apart from deep learning, where models are trained on many iterations through the data set until the loss function reaches a local minimum.  Whereas such a procedure makes sense for language engineering, it does not make sense for human learning: we don't relive the same exposure to data multiple times, and for healthy people, there is no point in learning after which performance degrades.  For instance, vocabulary learning is a continuous process straight into old age \citep{Keuleers:2015}. 

Note finally, that even though incremental learning is certainly superior for modeling realistic frequency effects, there are also cases where the end-state of learning can be the preferred choice of modeling. Incremental learning is much more computationally expensive which becomes a problem especially if the training set is large and frequencies are high. Moreover, in cases where simulated speakers are expected to have learned a phenomenon well enough, the end-state simulation might be sufficient.

In summary, the present modeling framework offers the possibility to approximate incremental human learning and the consequences of frequency of exposure for learning  in a cognitively motivated way \citep[see also][for learning in a multilingual setting]{Chuang:Bell:Banke:Baayen:2020}.

\subsection{Model complexity}

LDL mappings are costly in the number of connection weights, or equivalently, the number of beta coefficients.   For example, the mapping matrix $\mathbf{F}$ for the dataset discussed in Section~\ref{sec:semantic_roles} has 35 million weights (5913 $\times$ 5913 dimensions), rendering it much more costly in terms of the number of weights than deep-learning models, models such as AML, MBL, and recursive partitioning methods.  

\begin{figure}[tb!]
         \centering
         \includegraphics[width=\textwidth]{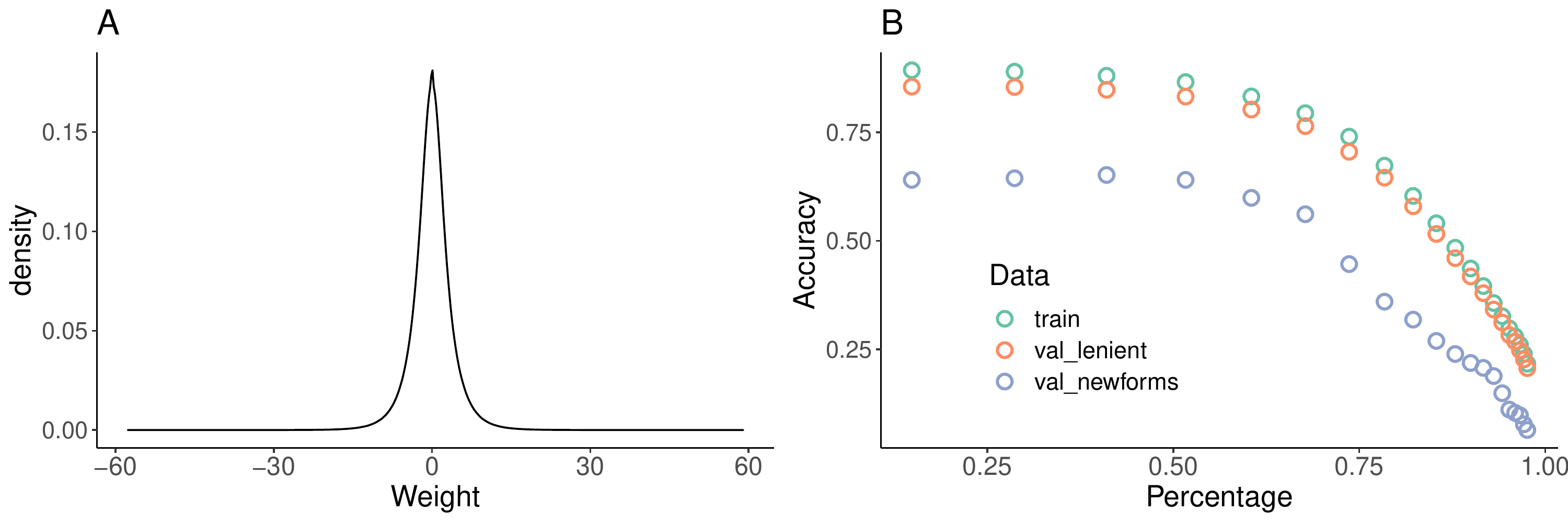}
    
    \caption{(A) Distribution of weights in the mapping matrix from form to meaning for the dataset with semantic roles. (B) Accuracy of the end-state model as a function of the proportion of connection weights close to zero are pruned. About 40\% of the weights can be set to zero without seriously affecting the performance of the model.}
    \label{fig:model_complexity}
\end{figure}

Inspection of the distribution of weights, however, clarifies that many weights are very close to zero.  Apparently, many  cues have low discriminative value. This suggests their connections can be pruned without seriously affecting model performance.  This can be tested by selecting a threshold $\vartheta$ and setting all absolute values in the mapping matrix that fall below this threshold to zero. Figure~\ref{fig:model_complexity} shows, for varying $\vartheta$, that up to 40\% of the small weights can be pruned without substantially impacting model performance with end-state of learning.  As neural pruning is part and parcel of human cortical development \citep[see, e.g.][]{gogtay2004dynamic}, an interesting topic for further research is to integrate incremental learning with neural pruning of uninformative connections.

\section{Discussion} \label{sec:discussion} 

In this study, we illustrated the methodological consequences of the many different choices that have to be made when modelling morphological systems within the discriminative lexicon framework, using LDL as modeling engine.  We illustrated these choices for the German noun system. This system is `degenerate', as many of its paradigm cells share the same word forms (homophones).  This system is also in many ways irregular: a noun's declension class can often not be fully predicted by its phonology, gender, or semantics \citep{kopcke1988schemas}. The results we obtained with LDL reflect this complexity. The model can learn word forms very well, achieving accuracies of more than 90\% on both comprehension and production when evaluated on training data. It can also generalize very well to new paradigm cells when it comes to word forms it has already seen, thanks to the ubiquitous homophony that characterizes German noun paradigms. However, it also mirrors the unpredictability of German inflections when it comes to word forms it has not seen before.  Accuracies for both comprehension and production suffer.  Nevertheless, the model shows some semi-productivity and succeeds in generalizing to many of the sub-regularities found in the German noun system \citep{wunderlich1999german},  reaching accuracies of 50\% on comprehension and 20\% on production.  Since German speakers encounter similar problems with new German word forms, as has been demonstrated in various wug studies \citep{zaretsky2013acquisition, mccurdy2020inflecting}, our model properly exhibits the limitations that are also characteristic for native speakers.  

In this study, we also probed the modeling of German nouns in context.  The rampant homophony that characterizes German noun paradigms is a straightforward consequence of considering words in isolation.   The amount of homophony can be substantially reduced by including articles,  in which case the model still performs well.  In context, case-inflected words typically do not realize a specific case meaning, but rather a specific semantic role.  As case endings typically do not stand in a one-to-one relation with semantic roles, we also examined to what extent we can make the model more realistic by replacing semantic vectors for cases with semantic vectors for a variety of semantic roles.  For the simulated dataset that we constructed, the model again performed well.  

For this dataset, we also demonstrated how the consequences of frequency of occurrence can be brought into the model, namely, by moving from the end-state of learning (estimated with regression) to incremental learning using the Widrow-Hoff learning rule. 

One limitation of the present approach is that most models have been using  very high-level abstract representations. The phone-based representation, for example, involves tremendous simplifications compared to for real speech, as variability in pronunciations is enormous \citep{Johnson:2004, Ernestus:Baayen:Schreuder:2002, ShafaeiBajestan:2021}. On the meaning side,  traditional case labels have no intrinsic semantic content, and although we can replace cases with semantic roles, these too are still too simplistic to be able to capture the full complexity of the semantics of words in context. However, we note that even with the present high-level representations, the model can still generate useful predictions.  We note here that various other studies carried out within this framework have successfully modeled a range of aspects of human lexical processing \citep[see][for further details]{Chuang:Baayen:2020}. In summary, even though the current framework undoubtedly misses out on a great number of nuanced but potentially informative features of forms and meanings in real language use, it can still serve as a useful linguistic tool to explore the strengths and weaknesses of morphological systems.

A question that inevitably arises in the context of computational modeling is how cognitively plausible a model is.  In the introduction, we called attention to the distinction made by \citet{breiman2001statistical} between statistical models and machine learning models.  We view LDL primarily as a statistical model that enables us to clarify, at a functional level of analysis, quantitative structure in the lexicon as well as understand the challenges a language processing system faces, without claiming that our model is cognitive reality. However, it is worth noting that LDL helps incorporate biologically and psychologically plausible learning into linguistic theory by making use of the principle of error-driven learning (when training the model incrementally). The very simple learning rules of Widrow-Hoff and Rescorla-Wagner have been shown to excellently explain phenomena from a range of domains in e.g. biology and psychology \citep[see, e.g.,][]{Rescorla:1988,Schultz:2008,Marsolek:2008,oppenheim2010dark,Trimmer:2012}.

It is possible to take the model as point of departure for addressing questions at the level of neural organization in the brain.   For instance, \citet{Heitmeier:Baayen:2020} were interested in clarifying whether the framework of the discriminative lexicon properly predicts the dissociations of form and meaning observed for aphasic speakers producing English regular and irregular past-tense forms, following \citet{Joanisse:Seidenberg:1999}. They took the unordered banks of units of form and meaning (the column dimensions of the $\bm{C}$ and $\bm{S}$ matrices) and projected them onto two-dimensional surfaces approximating, however crudely, cortical maps.  This made it possible to lesion the network in a topologically cohesive way, rather than by randomly taking out connections across the whole network.  For projection, they made use of an algorithm from physics (\url{http://www.schmuhl.org/graphopt/}) for displaying graphs, but temporal self-organizing maps   \citep[TSOMs,][]{ferro2011self,chersi2014topological} offer a much more fine-grained and principled way for modeling morphological organization that builds on principles of error-driven learning.

Deep learning algorithms provide the analyst with powerful modeling tools, but it seems  they are too powerful \citep[see, e.g.,][]{mccurdy2020inflecting} for understanding not only the strengths but also the weaknesses and the frailties of human lexical memory and lexical processing.  However, linguistic models are in a different way also too powerful on the one hand, and too underspecified on the other hand.  Paradigms are typically constructed to accommodate any contrast between forms and inflectional functions, even when a contrast is attested only for a few forms in the language.  The result is an overabundance of homophones, which are severely underspecified with respect to their real meanings in actual language use (such as their semantic roles).  Furthermore, in actual language use, inflected forms can occur at very different frequencies and some are never encountered at all \citep{karlsson1986frequency, janda2018less}, which in turn has demonstrable consequences for lexical processing \citep{Loo:2018a}.\footnote{Note that we do not claim that rare inflected word forms cannot be processed. Generally, the more regular a morphological system, the more easily the model can predict new forms \citep[e.g. in Estonian,][]{chuang2020estonian}, while in semi-productive cases such as German or Maltese \citep{nieder2021comprehension} generalisation is much more difficult.}  An interesting challenge for further research is to clarify how different degrees of paradigm economy \citep{ackerman2013morphological} are reflected in the matrices that define mappings between form and meaning within the framework of the discriminative lexicon.  

In this study, we have provided an overview of the many choice points that arise in modeling with LDL, each of which requires knowledge of morphology and morphological theory.   The implications of our approach to psycho-computational modeling for morphological theory depends on the specifics of a given specific theory of morphology.  Our approach is broadly consistent with usage-based approaches to morphology \citep{Bybee:85,Bybee:2010}, and with Word and Paradigm Morphology \citep{Blevins:2016}.  It is less clear whether our modeling approach is informative for theories that are only interested in defining possible words.  With this methodological study, we have shed some light on the many questions and issues that do not arise in formal theories of morphology, but that have to be addressed in a linguistically informed way when the goal of one's theory is to better understand, and predict, in all its complexity, human lexical processing across comprehension and production.  

\section{Acknowledgments}
This work was supported by ERC-WIDE (European Research Council—Wide Incremental learning with Discrimination nEtworks), grant number 742545. We acknowledge support by Open Access Publishing Fund of University of Tübingen. We thank Ruben van de Vijver and the two reviewers for their helpful comments on earlier versions of this manuscript.

\bibliography{bib2}

\end{document}